\documentclass{article}
\usepackage[utf8]{inputenc}

\usepackage{etoolbox}
\newcommand{\arx}[1]{\iftoggle{sts}{}{#1}}
\newcommand{\sts}[1]{\iftoggle{sts}{#1}{}}
\newtoggle{sts}
\global\togglefalse{sts}

\RequirePackage{amsmath,amsfonts,amssymb} 

\usepackage{url}
\usepackage{algorithm, algpseudocode}

\usepackage{amsmath, amssymb} 
\usepackage[shortlabels]{enumitem}
\sts{\usepackage{color}}
\usepackage{multirow}
\usepackage{xspace}
\usepackage{rotating}
\usepackage{graphicx}
\usepackage{caption}
\sts{\usepackage{xcolor}}
\usepackage[normalem]{ulem}
\usepackage{tikz}

\usepackage{nicefrac}
\usepackage{mathtools}
\mathtoolsset{showonlyrefs,showmanualtags}

\newtheorem{theorem}{\textbf{Theorem}}
\newtheorem{lemma}[theorem]{\textbf{Lemma}}

\newtheorem{proposition}[theorem]{Proposition}

\newtheorem{example}{Example}

\newtheorem{assumption}{Assumption}

\renewcommand{\cite}{\citep}

\newcommand{\EE}{\mathbb{E}}

\newcommand{\Indi}{\mathbb{I}}

\DeclareMathOperator*{\argmin}{arg\,min}
\DeclareMathOperator*{\argmax}{arg\,max}

\newcommand{\Fcal}{\mathcal{F}}

\newcommand{\Wcal}{\mathcal{W}}
\newcommand{\Vcal}{\mathcal{V}}
\newcommand{\Gcal}{\mathcal{G}}
\newcommand{\Hcal}{\mathcal{H}}

\newcommand{\Mcal}{\mathcal{M}}
\newcommand{\Scal}{\mathcal{S}}

\newcommand{\Acal}{\mathcal{A}}
\newcommand{\Tcal}{\mathcal{T}}

\newcommand{\Ecal}{\mathcal{E}}
\newcommand{\Lcal}{\mathcal{L}}

\newcommand{\Rmax}{R_{\max}}

\newcommand{\Vmax}{V_{\max}}
\newcommand{\RR}{\mathbb{R}}
\newcommand{\Xcal}{\mathcal{X}}

\newcommand{\Dcal}{\mathcal{D}}
\newcommand{\CA}{{C_{\mathcal{A}}}}
\newcommand{\Ehat}{\widehat{\Ecal}}
\newcommand{\Lhat}{\widehat{\Lcal}}
\newcommand{\dm}{\mathsf{d}}
\newcommand{\VS}[1]{\Fcal_{\epsilon_0}^{#1}}
\newcommand{\picomp}{\pi_{\textrm{cp}}}
\newcommand{\Cavg}{C^{\textrm{avg}}}
\newcommand{\Csq}{C^{\textrm{sq}}}
\newcommand{\Phat}{\widehat{P}}
\newcommand{\Pcal}{\mathcal{P}}
\newcommand{\pirb}{\pi_{\textrm{rb}}^\star}
\newcommand{\piref}{\pi_{\textrm{ref}}}
\newcommand{\eff}{\textrm{eff}}
\newcommand{\proj}{\textrm{Proj}}

\newcommand{\para}[1]{\vspace{0.4em} \noindent \textbf{#1}~}
\usepackage{mdframed}
\mdfsetup{nobreak=true}

\usepackage{natbib}
\setcitestyle{authoryear,round,semicolon}

\usepackage[svgnames]{xcolor}
\colorlet{txblue}{RoyalBlue!70!NavyBlue}
\usepackage{xurl}

\usepackage{hyperref}
\hypersetup{colorlinks=true,
            linkcolor=txblue,  
            allcolors=txblue,
            urlcolor=black,
            breaklinks=true,
            linktoc=all
}

\usepackage{geometry}
\usepackage{parskip}

\let\oldparagraph\paragraph
\renewcommand{\paragraph}[1]{\oldparagraph{#1.}}

\usepackage[T1]{fontenc}
\usepackage[scaled=.91]{helvet}
\usepackage{inconsolata}
\usepackage{microtype}

\newcommand{\keywords}[1]{\begin{quotation}\noindent\textbf{Key words and phrases:} #1\end{quotation}}

\title{Offline Reinforcement Learning in Large State Spaces:\\ Algorithms and Guarantees}

\author{Nan Jiang \and Tengyang Xie}

\date{}

\newcommand{\printaddresses}{
  \begingroup
  \renewcommand\thefootnote{}
  \footnote{\it To appear in Statistical Science.}
  \footnote{\it Nan Jiang is Associate Professor of Computer Science, University of Illinois Urbana-Champaign (email: \href{mailto:nanjiang@illinois.edu}{\tt nanjiang@illinois.edu}). Tengyang Xie is Assistant Professor of Computer Science, University of Wisconsin--Madison (email: \href{mailto:tx@cs.wisc.edu}{\tt tx@cs.wisc.edu}).}
  \addtocounter{footnote}{-2}
  \endgroup
}

\begin{document}

\maketitle

\printaddresses

\begin{abstract}
This article introduces the theory of offline reinforcement learning in large state spaces, where good policies are learned from historical data without online interactions with the environment. Key concepts introduced include expressivity assumptions on function approximation (e.g., Bellman completeness vs.~realizability) and data coverage (e.g., all-policy vs.~single-policy coverage). A rich landscape of algorithms and results is described, 
depending on the assumptions one is willing to make and the   sample and computational complexity guarantees  one wishes to achieve. We also discuss open questions and connections to adjacent areas. 

\end{abstract}


\section{Introduction}
Reinforcement learning (RL), despite the word ``learning'' in its name, has been historically about using sampling-based methods for \textit{computational tasks}. As Sutton once put it:\footnote{\url{http://incompleteideas.net/RL-FAQ.html}.} 

\begin{center} \it
``Much of the [RL] field does not concern learning at all, but just planning\arx{\\} from \ldots a model of the environment.''
\end{center}

This is indeed largely the case in empirical (deep) RL research, where algorithms find near-optimal policies by interacting with simulation environments to sample data trajectories. The goal here is very clear: finding a good policy using a given amount of computation, which includes both the cost of the algorithm and that of sampling data from the simulator. 

While this paradigm has led to impressive successes in difficult simulation tasks \citep{silver2017mastering,vinyals2019grandmaster}, it becomes increasingly clear that the above paradigm is insufficient for many potential applications we hope to apply RL to, including adaptive clinical trials \citep{zhao2009reinforcement,zhao2011reinforcement,tang2021model,shiranthika2022supervised}, recommendation systems and customer relationship management \citep{zheng2018drn,afsar2022reinforcement}, online education \citep{bassen2020reinforcement,nie2023understanding}, and more. A commonality of the above scenarios is that human patients/users/students are part of the ``environment'', and it can be  difficult to come up with accurate simulators for the psychological/biological aspects of humans. Therefore, the only environment we have access to is the \textit{real} one, i.e., the real patients, the real users, etc. 

This leads to a further problem. Most simulation-based RL algorithms are \textit{online}:\footnote{Generally, online RL can handle more general and real-world tasks beyond simulations when interactions are low-stakes and cheap.} while interacting with the environments to collect data, the algorithms will experiment with whatever decisions they deem suitable and observe their effects. These decisions can lead to undesirable outcomes, especially at the early stages of learning when the algorithm has little knowledge of the environment. This is not a problem when the environment is a simulator,  but can lead to serious consequences when human users/customers/patients are part of the environment. 
Offline RL, which learns from pre-collected data without online interactions, is an answer to this challenge. For real-world environments, such data can come from logging the normal operations of the system without changing how decisions are made. 

This article aims to provide a brief introduction to the core concepts and ideas in   offline RL theory, focusing on the following two aspects:
\begin{itemize}[leftmargin=*]
\item \textit{\textbf{Data: }} As we will see, the restriction of no online interactions brings serious algorithmic challenges, and learning guarantees are largely subject to the quality and the quantity of the dataset, requiring us to elevate data as the first and foremost consideration. Hence, our discussion will be focused on the \textit{statistical} aspects of offline RL, mostly the sample complexities, with occasional remarks about computational efficiency. \vspace*{.5em}
\item \textit{\textbf{Function approximation: }} Historically, RL theory with complexity guarantees started in settings where the number of states is finite and small, known as ``tabular'' RL \citep{kearns2002near}. In this article we will skip tabular RL and directly address large state spaces where tractable learning often requires function approximation. Readers may find it surprising that, unlike standard settings in other areas of machine learning and statistics, \textit{a hypothesis class that perfectly captures a target function (a.k.a.~\emph{realizability}) is often \textbf{insufficient} for learning the said function in RL}. This fact has deep implications in not only learning but also evaluation and testing, and the literature considers a rich variety of function-approximation assumptions that lead to different algorithms and guarantees. 
\end{itemize}

\section{A Gentle Start: The Curse of Horizon}
Offline RL promotes a \emph{data}-driven paradigm similar to supervised learning (SL). In this section, we quickly review a minimal theoretical setup of SL and establish its counterpart in offline RL using importance sampling. This analogy will help us appreciate the unique challenges RL faces and prepare for the subsequent discussion of alternative methods. 

To start, consider a standard supervised classification setup (the notation will only be used for making the comparison between SL and RL and will not be carried to subsequent sections): we have dataset $\{(X_i, Y_i)\}_{i=1}^n$ drawn i.i.d.~from some distribution, where $X_i \in \Xcal$ is the input features (say images) and $Y_i \in \{-1, 1\}$ is the binary label (say whether the image contains a cat), and the goal is to learn $h: \Xcal\to\{-1, 1\}$ that makes accurate predictions of $Y$ from $X$, measured by the error rate: $\mathrm{err}(h) \coloneqq \EE[\Indi[Y_i \ne h(X_i)]].$ 

A key but perhaps under-appreciated fact is that \emph{$\mathrm{err}(h)$ can be efficiently estimated from data for a fixed $h$} by 
$\widehat{\mathrm{err}}(h) \coloneqq \tfrac{1}{n}\sum_{i=1}^n \Indi[Y_i \ne h(X_i)].$ 
This is because $\widehat{\mathrm{err}}(h)$ is the average of i.i.d.~r.v.s $\{\Indi[Y_i \ne h(X_i)]\}_{i=1}^n$, and Hoeffding's inequality tells us that the deviation from true mean is bounded as
\begin{align} \label{eq:sl_eval}
\sts{\textstyle} |\widehat{\mathrm{err}}(h) - \mathrm{err}(h)| \le \sqrt{\frac{1}{2n}\log \frac{2}{\delta}} 
\end{align}
with probability at least $1-\delta$. 

Eq.\eqref{eq:sl_eval} is the foundation of everything in SL: we train by Empirical Risk Minimization (ERM) $\argmin_{h\in\Hcal} \widehat{\textrm{err}}(h)$, and a basic generalization error bound that depends on $\log|\Hcal|$ can be obtained from Eq.\eqref{eq:sl_eval} and union bounding over $\Hcal$. In practice, we may try different training algorithms, and rely on $\widehat{\mathrm{err}}(h)$ estimated from holdout datasets for model selection and evaluation (test). 

\subsection{Off-Policy Evaluation by IS} \label{sec:IS}
To establish the counterpart of ERM-then-evaluate scheme for offline RL, the central question is \emph{how to estimate the performance of a candidate solution and establish a  guarantee similar to Eq.\eqref{eq:sl_eval}.} In RL, the learning algorithms output decision-making strategies, or \emph{policies}, which are generally different from the policies used to collect the dataset in the first place. Evaluating a policy given data collected by a different policy is known as the problem of \textit{off-policy evaluation} (OPE). 

To introduce OPE methods, we consider a standard setup for RL in infinite-horizon discounted Markov Decision Processes (MDPs): An MDP $(\Scal, \Acal, P, R, \gamma, d_0)$ is specified by its state space $\Scal$, action space $\Acal$, transition dynamics $P:\Scal\times\Acal\to\Delta(\Scal)$, reward function $R: \Scal\times\Acal\to[0, \Rmax]$, discount factor $\gamma \in [0, 1)$, and initial distribution $d_0$. Here rewards are deterministic and non-negative, which are inconsequential simplifications. 

We also assume $\Scal$ and $\Acal$ are discrete and finite for convenience; the results we introduce scale to arbitrarily large $\Scal$ (and large $\Acal$ sometimes), and can be adapted to continuous $\Scal$ under appropriate measure-theoretic notation.\footnote{In fact, some early work did consider continuous state spaces \citep{antos2008learning}, but later works often choose finite spaces for readability \citep{chen2019information}.} A (stationary and stochastic) policy $\pi: \Scal\to\Delta(\Acal)$ is a decision-making strategy, and induces a distribution over \textit{trajectories}:
$$
\tau \coloneqq (s_0, a_0, r_0, s_1, a_1, r_1, \ldots, s_{H-1}, a_{H-1}, r_{H-1}, \ldots),
$$
described by the following generative process: $s_0 \sim d_0$, $a_t \sim \pi(\cdot \mid s_t)$, $r_t = R(s_t, a_t)$, $s_{t+1} \sim P(\cdot \mid s_t, a_t)$, $\forall t\ge 0$. The sampling process can go on forever ($t\to\infty$), but for simplicity we assume that after at most $H$ steps the process always goes into a self-loop state with $0$ reward (a.k.a.~an \textit{absorbing state}) which marks the termination of the system (patient completing a multi-stage treatment program, user concluding a multi-round conversation with a chatbot, etc.).\footnote{This is only needed by importance sampling. When it does not hold, one can truncate an infinitely long trajectory at an \textit{effective horizon} $H = O(1/(1-\gamma))$ due to discounting, as rewards after $H$ steps are discounted so heavily that omitting them only incurs a small error.}

Given the MDP model, it suffices to specify two things to define an estimation problem: 

\para{Estimand $J(\pi)$:} Given a policy $\pi$ we want to evaluate (often called a target/evaluation policy), the estimand is the expected discounted return, defined as:
$$ \sts{\textstyle} 
J(\pi)\coloneqq \EE_{\pi}\arx{\left}[\sum_{t=0}^\infty \gamma^{t} r_t\arx{\right}], 
$$
where $\EE_{\pi}[\cdot]$ refers to distribution of trajectories under policy $\pi$. We will also use $\Pr_{\pi}[\cdot]$ to refer to probabilities under the same distribution. 
This is a standard objective that measures how much total reward a policy is able to collect in expectation. When we assume that the process terminates in $H$ steps, $\sum_{t=0}^\infty$ can be replaced by $\sum_{t=0}^{H-1}$ since all rewards after $t=H$ are $0$. 

\para{Dataset:} In OPE, we have data trajectories (or episodes) collected from a different policy, 
$\pi_D$, often referred to as the behavior/logging policy. More concretely, the dataset is 
$
\{ \tau^{(i)}\coloneqq(s_0^{(i)}, a_0^{(i)}, r_0^{(i)}, \ldots, s_{H-1}^{(i)}, a_{H-1}^{(i)}, r_{H-1}^{(i)}) \}_{i=1}^n,
$ where all actions $a_t^{(i)} \sim \pi_D(\cdot \mid s_t^{(i)})$. 

\para{IS Estimator:} The importance sampling (IS) estimator \citep{precup2000eligibility}, also known under the names of importance weighting and inverse propensity score (IPS), forms an unbiased estimate of $J(\pi)$ using a single trajectory  (the $(i)$ subscript is omitted):
\begin{align} \label{eq:IS} 
\textrm{IS}(\tau) \coloneqq \left(\prod_{t=0}^{H-1} \frac{\pi(a_t \mid s_t)}{\pi_D(a_t \mid s_t)}\right)\left(\sum_{t=0}^{H-1} \gamma^t r_t\right).
\end{align}
One can show that, as long as $\pi_D(a \mid s)>0$ for all $(s,a)$ where $\pi(a \mid s) > 0$ (or informally,  $\pi/\pi_D < \infty$), 
$$\EE_{\pi_D}[\textrm{IS}(\tau)] = \sts{\textstyle} \EE_{\pi}\arx{\left}[\sum_{t=0}^\infty \gamma^t r_t\arx{\right}] \equiv J(\pi),$$ 
because the importance weight $\prod_{t=0}^{H-1} \frac{\pi(a_t \mid s_t)}{\pi_D(a_t \mid s_t)} = \frac{\Pr_\pi[\tau]}{\Pr_{\pi_D}[\tau]}$ converts the distribution from being sampled with $\pi_D$ to $\pi$ in expectation. $\pi/\pi_D < \infty$ is a \textit{coverage} condition. In problems with small action spaces, this can be satisfied for any target policy $\pi$ when $\pi_D$ is properly randomized and puts nontrivial probabilities on all actions.

Then, given $n$ trajectories, it is straightforward to form an estimation of $J(\pi)$ by $\widehat J_{\textrm{IS}}(\pi) = \frac{1}{n}  \sum_{i=1}^n \textrm{IS}(\tau^{(i)}).$ 
$\textrm{IS}(\tau)$ in Eq.\eqref{eq:IS} is the most basic version of IS and can be improved in many ways, such as using a data-dependent normalization factor (``weighted IS'') and  control variates (``doubly robust'') \citep{jiang2016doubly, thomas2016data}. Nevertheless, these estimators share similar high-level sample-complexity characteristics as we discuss below.

\para{Guarantee:} Since $\widehat J_{\textrm{IS}}(\pi)$ is the average of i.i.d.~r.v.s $\textrm{IS}(\tau^{(i)})$, we immediately have an estimation guarantee similar to Eq.\eqref{eq:sl_eval} for SL, except for one thing: Hoeffding's inequality depends on the range of the r.v.s., which is $[0, 1]$ for Eq.\eqref{eq:sl_eval}. For IS,  $\sum_{t=0}^{H-1} \gamma^t r_t \in [0, \Vmax]$ where $\Vmax \coloneqq \Rmax/(1-\gamma)$ is a standard range parameter for the total reward in MDPs, but we also need a bound on the importance weight $\prod_{t=0}^{H-1} \frac{\pi(a_t \mid s_t)}{\pi_D(a_t \mid s_t)}$. Assuming 
$$\max_{s,a} \frac{\pi(a \mid s)}{\pi_D(a \mid s)} \le \CA, 
$$
it immediately follows that $\prod_{t=0}^{H-1} \frac{\pi(a_t \mid s_t)}{\pi_D(a_t \mid s_t)} \le (\CA)^H$. Then, using a standard concentration argument,\footnote{Here Bernstein's inequality is used to leverage the ``low variance'' (compared to its range) of importance weights: given any non-negative function $\rho$ such that $\EE[\rho] = 1$ (which is satisfied by importance weights in general), we always have $\EE[\rho^2] \le \|\rho\|_\infty$, i.e., variance scales with range linearly instead of quadratically. \label{ft:is-var}} we have
\begin{align} \label{eq:IS_bound}
\sts{\textstyle} |\widehat J_{\textrm{IS}}(\pi) - J(\pi)| \lesssim \Vmax \sqrt{\frac{(\CA)^H}{n}\log \frac{1}{\delta}},
\end{align}
where ``$\lesssim$'' means LHS$= O(\textrm{RHS})$ and suppresses multiplicative absolute constants. 

With the IS estimator, we can establish an exact parallel of SL's framework for offline RL: given a class of policies $\Pi$ we wish to optimize, we can simply perform 
$$
\argmax_{\pi\in \Pi} \widehat J_{\textrm{IS}}(\pi)
$$
to approximately find the best policy in $\Pi$. The learned policy can then be evaluated, again using IS but on a holdout dataset, for model selection and testing.

\subsection{The Curse of Horizon} \label{sec:curse}
The exact parallel between the IS-based framework and SL brings many desirable properties (which we will certainly miss in subsequent sections!). In fact, the framework is used in industrial applications of contextual bandits, which can be viewed as 1-step RL ($H=1$) \citep{li2010contextual}.

Unfortunately, the framework has a crucial caveat for multi-step RL: the exponential term $(\CA)^H$, which enters the sample complexity of \emph{even evaluating a single policy} (Eq.\eqref{eq:IS_bound})! 
This term can be small if $\CA$ is close to $1$. In fact, if $\CA = 1 + O(1/H)$,   $(\CA)^H$ will be a constant. However, that restricts us  to only target policies that are very close to the behavior policy. 

To see why this can be unsatisfactory, 
consider an MDP with only one state and two actions. The state always transitions back to itself. Behavior policy collects data using a uniformly random policy. Intuitively, even with a moderately large dataset  we will have the needed information to evaluate any policy accurately, but IS still suffers exponential variance when evaluating a deterministic policy. 
This clearly contradicts the intuition that this is a simple problem and should not require an exponential-in-horizon dataset for learning. 
The rest of this article will be largely focused on how to address such a ``curse of horizon''. 

Some final remarks about IS before we move on:

\para{On $\CA$ and $|\Acal|$} If we have full control over the choice $\pi_b$ but do not have a priori information about $\pi$, the best-case scenario for $\CA$ is $|\Acal|$ when $\pi_b$ chooses actions  uniformly randomly (a.k.a.~uniform exploration). In contextual bandit applications \citep{li2010contextual}, a small amount of uniform exploration is often injected into the system to ensure that the collected data can be useful for IS. On the other hand, $\CA$ can be large or even infinite if the behavior policy lacks randomization. 

Moreover, IS can also work for large action spaces. For stochastic $\pi$ and $\pi_D$, it is possible  that $\pi/\pi_D$ is well bounded  
even if $|\Acal|$ is large or even infinite (e.g., Gaussian policies in robotics \citep{mandlekar2022matters}). In fact, popular deep RL algorithms such as PPO \citep{schulman2017proximal} use one-step importance sampling and are frequently applied to problems with large action spaces.

\para{``Constants'' in Finite-sample Guarantees} Eq.\eqref{eq:IS_bound} translates to a sample complexity guarantee of 
$$
n = O((\CA)^H \Vmax^2 \log\tfrac{1}{\delta} / \epsilon^2),
$$ 
for estimating $J(\pi)$ to $\epsilon$ accuracy with high probability (i.e., at least $1-\delta$). 
This is also a good example showing the importance of \emph{finite-sample results} in RL, as asymptotic identification would ``hide away'' the exponential term $(\CA)^H$. For the same reason, the convention of only highlighting the dependence on $\epsilon$ (or the number of interaction rounds $T$ in online RL setups) and treating all other quantities as ``constants'', which is common in adjacent fields such as online learning, can also be inappropriate. 

As a bonus, discrete vs.~continuous $\Scal$ and $\Acal$ are qualitatively different for asymptotic identification,\footnote{With discrete $\Scal$ and $\Acal$, asymptotically we can visit every $(s,a)$ pair infinitely often given proper exploration and infinite data, which is impossible for continuous $\Scal$ and $\Acal$.} but their boundary is blurred for finite-sample results. If a finite-sample guarantee has no dependence on $|\Scal|$, extending it to continuous $\Scal$ is mostly a matter of formality. This allows us to adopt a minimal setup of finite spaces, while the algorithms and insights are directly applicable to continuous spaces.

\section{Value Function Estimation} \label{sec:vf}
Back to the 1-state-2-action MDP example: what makes it feel tractable? The answer is Markovianity: despite there exist exponentially many action sequences, they all pass through the same \textit{state}. 
As we will see, methods that properly leverage Markovianity enjoy guarantees when  states and actions visited by the target policy are \textit{covered}---in a precise technical sense explained later---in the data.  In contrast, IS completely ignores the notion of state and essentially considers an exponential tree of ever-branching trajectories, requiring trajectories generated by the target policy to be covered.   (This makes IS directly applicable to partially observed domains, which we will discuss at the end.)

For problems with finite and small state spaces, one can certainly estimate the transition and reward for each state-action pair separately and compute the policy's return in the estimated MDP, known as (tabular) ``certainty equivalence'' \citep{kumar2015stochastic}.\footnote{Tabular certainty equivalence is a special case of FQE/BRM when the function classes contain all possible functions over $\Scal\times\Acal$, and hence the analyses of FQE/BRM are directly applicable.} 
The question is how to leverage Markovianity in a way that scales to large state spaces. For that we must turn to a familiar object: \textit{value functions}.

\para{Value Functions} Given $\pi: \Scal\to\Delta(\Acal)$, its (Q-)value function is 
$$
Q^\pi(s,a) \coloneqq \sts{\textstyle} \EE_{\pi}\arx{\left}[\sum_{t=0}^\infty \gamma^t r_t \,|\, s_0 = s, a_0 = a\arx{\right}],
$$
which is the expected return when a trajectory  starts with $(s,a)$ and actions follow $\pi$ from $t=1$ onwards.\footnote{``$s_0=s, a_0=a$'' are better thought of as defining a new distribution of trajectories, but we will follow the convention and write them as conditions.} Once $Q^\pi$ is known, $J(\pi)$ can be extracted as 
$$J(\pi) = \EE_{s \sim d_0}[Q^\pi(s,\pi)],$$ 
where $f(s,\pi)$ is the shorthand for $\EE_{a\sim \pi(\cdot \mid s)}[f(s, a)]$. $\EE_{d_0}[\cdot]$ can be easily estimated from an i.i.d.~bag of states sampled from $d_0$ (e.g., $\{s_0^{(i)}\}_{i=1}^n$ from Section~\ref{sec:IS}), and we   assume $d_0$ is known to simplify presentation. 

We will show below that learning Q-functions is an effective approach to overcoming the curse of horizon. There are two key questions: \vspace{.35em}

\noindent \textbf{Q1:} How to estimate $Q^\pi$ from data? \\
\noindent \textbf{Q2:} What guarantees can we say about estimated $J(\pi)$?
\vspace{.35em}

We start with \textbf{Q1} and introduce two important ideas for estimating $Q^\pi$ in Sections~\ref{sec:fqe} and \ref{sec:brm}, and address \textbf{Q2} by providing analyses based on a notion of state-action coverage in Section~\ref{sec:error-prop}.

\subsection{Fitted-Q and Bellman Completeness} \label{sec:fqe} Estimation of $Q^\pi$ is typically based on the fact that it is the unique fixed point of the (policy-specific) Bellman operator, i.e., 
\begin{align} \label{eq:bellman}
    Q^\pi = \Tcal^\pi Q^\pi,
\end{align} 
where $\Tcal^\pi: \RR^{\Scal\times\Acal} \to \RR^{\Scal\times\Acal}$ is defined as: $\forall f\in \RR^{\Scal\times\Acal}$, 
\begin{align} \label{eq:Tpi}
(\Tcal^\pi f)(s,a)\coloneqq \EE_{r = R(s,a), s'\sim P(\cdot \mid s,a)}[r + \gamma f(s', \pi)].
\end{align}
Computational algorithms for solving $Q^\pi$ are often based on dynamic programming (DP): for example, value iteration (VI) repeatedly applies $\Tcal^\pi$ to an arbitrary initial function $f_0$, and the contraction property\footnote{$\forall f, f'\in\RR^{\Scal\times\Acal}$, $\|\Tcal^\pi f - \Tcal^\pi f'\|_\infty \le \gamma \|f - f'\|_\infty$.} implies that $(\Tcal^\pi)^n f_0 \to Q^\pi$ with a geometric convergence speed in $\|\cdot\|_\infty$, that is,
\begin{align} \label{eq:contraction}
\|(\Tcal^\pi)^n f_0 - Q^\pi\|_\infty \le \gamma^n \| f_0 - Q^\pi \|_\infty.
\end{align}
This motivates one of the most popular family of algorithms for value-function estimation, which approximates the operator $\Tcal^\pi$ from data. Note that $\Tcal^\pi f$ in Eq.\eqref{eq:Tpi} takes the form of a conditional expectation, which can be written as a regression problem:  
\begin{align} \label{eq:regression}
\Tcal^\pi f  \in \argmin_{f' \in \RR^{\Scal\times\Acal}} \Lcal(f';f, \pi),
\end{align}
where $\Lcal(f';f, \pi) \coloneqq\EE_D[(f'(s,a) - r - \gamma f(s',\pi))^2].$ This expression suggests that we may approximate $\Tcal^\pi$ with a least-square regression, and for that we must first clarify the form of data we use in this section. 

\para{Data Protocol} Here $D$ is a shorthand for the distribution of $(s,a,r,s')$ observed in the offline data. In the rest of this article, we no longer need trajectory data, and learn with these  transition tuples which can be extracted as $\tau^{(i)} \to (s_0^{(i)}, a_0^{(i)}, r_0^{(i)}, s_1^{(i)}), (s_1^{(i)}, a_1^{(i)}, r_1^{(i)}, s_2^{(i)}), \ldots$. In such a case, the concentration argument for estimation needs to take into account the dependencies between tuples from the same trajectory \citep{antos2008learning}. We instead consider a standard simplification, that the dataset $\Dcal$ consists of i.i.d.~transition tuples: $(s,a,r,s') \sim D \Leftrightarrow$
$$
(s,a)\sim d^D, r = R(s,a), s' \sim P(\cdot \mid s,a).
$$

\para{Fitted-Q Algorithm} We are now ready to describe Fitted-Q  Evaluation (FQE), which can be viewed as the prototype or theoretical version of many empirically popular methods, including the TD family \citep{mnih2015human}. Fitted-Q assumes a function class $\Fcal$ for modelling $Q^\pi$ and chooses an arbitrary initialization $f_0 \in \Fcal$. Then, it solves a sequence of least-square regression algorithms: 
\begin{mdframed}
\begin{align} \label{eq:FQE}
f_k \gets \argmin_{f'\in\Fcal} \Lhat(f';f_{k-1},\pi),
\end{align}
\end{mdframed}
where $\Lhat$ is the empirical approximation of $\Lcal$ based on dataset $\Dcal$  (we will omit the formula for such straightforward estimations henceforward): 
$$
\sts{\textstyle} \Lhat(f';f,\pi) \coloneqq \frac{1}{|\Dcal|} \sum_{(s,a,r,s') \in \Dcal} (f'(s,a) - r - \gamma f(s',\pi))^2.
$$ 

\para{Divergence of Fitted-Q} Before we can analyze the algorithm, there is a very serious problem: Fitted-Q can diverge under arguably very strong assumptions.

\begin{proposition}[\sts{\citep{tsitsiklis1996feature}}\arx{\citealp{tsitsiklis1996feature}}] \label{prop:deadly}
FQE can diverge even when all of the following hold:
\begin{enumerate}
\item $|\Dcal|=\infty$ and  minimization in Eq.\eqref{eq:FQE} is exact.
\item $\Fcal$ is a 1-dim linear function class that exactly captures $Q^\pi$, i.e., $Q^\pi \in \Fcal$ (a.k.a.~\emph{realizability}).
\end{enumerate}
\end{proposition}
The divergence happens under seemingly perfect assumptions: infinite data, exact optimization, simple function class, and exact realizability. Moreover, this so-called ``deadly triad'' phenomenon \citep{van2018deep} is not merely a theoretical construction: deep RL algorithms are known for their instability and training divergence is commonly observed \citep{wang2021instabilities}. So what is going wrong?

The problem is that FQE solves a \textit{sequence} of regression problems, $(s,a) \mapsto r + \gamma f_{k-1}$, where the regression target (also known as the TD target) depends on $f_{k-1}$, the function from the last iteration. Therefore, for FQE to closely mimic VI, we need \emph{every regression in this sequence to be well-specified}, i.e., $\Fcal$ must include (a close approximation of) the Bayes-optimal predictor, $\Tcal^\pi f_{k-1}$, for every $k$. Given that $f_{k-1}$ depends on data randomness, we can relax $f_{k-1}$ to any function in $\Fcal$ to obtain an assumption independent of the data randomness, leading to  a very  important assumption in modern offline RL theory and the major expressivity assumption of this section:

\begin{mdframed}
\begin{assumption}[\textbf{Bellman Completeness} (for $\Tcal^\pi$)] \label{asm:complete}
$$\Tcal^\pi f \in \Fcal, ~ \forall f\in \Fcal.$$
\end{assumption}
\end{mdframed}

\begin{figure*}[t]
\centering
\includegraphics[width=.9\textwidth]{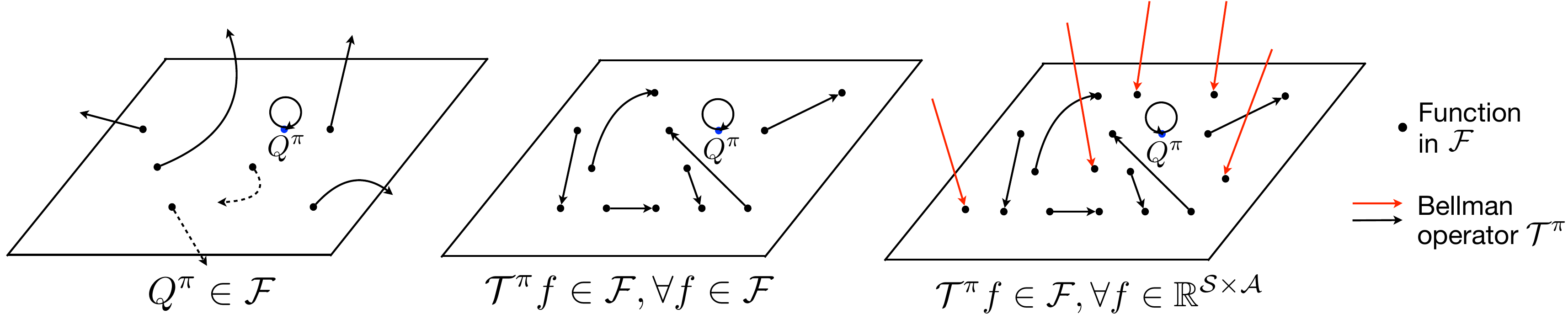}
\caption{Figurative illustration of different expressivity assumptions on the value-function class $\Fcal$. \textbf{Left:} Only realizability $Q^\pi \in \Fcal$ is assumed, and Bellman operator (``$\to$'') can generally take   functions in $\Fcal$ out of the class. \textbf{Middle:} Bellman-completeness, where $\Fcal$ is closed under $\Tcal^\pi$. \textbf{Right:} All functions, including those not in $\Fcal$, have their image in $\Fcal$ (Section~\ref{sec:pevi}).    \label{fig:complete}}
\end{figure*}

The assumption asserts that the function space is \textit{closed}
under the operator $\Tcal^\pi$, and is thus also referred to as Bellman closure/closedness; see Figure~\ref{fig:complete} for an illustration. 
For finite $\Fcal$, completeness implies realizability ($Q^\pi = (\Tcal^\pi)^\infty f \in \Fcal$) and is generally a stronger assumption. 
The very nature of the assumption is still a topic of debate; below we present several different angles. 

\para{Information-theoretic Angle}
If we know that the true MDP is in some MDP class $\Mcal$ with bounded log-cardinality, then an (approximately) Bellman-complete $\Fcal$ with a mild blow-up in size can be constructed by repeating $\Fcal \gets \Fcal \bigcup \{\Tcal^\pi_{M} f: f\in\Fcal, M \in \Mcal\}$ $O(H)$ times, where $H$ is the effective horizon (Section~\ref{sec:IS}) and the subscript $M$ in $\Tcal_M^\pi$ means the Bellman operator is defined w.r.t.~the MDP $M$.\footnote{The construction is more straightforward for the $(\Fcal, \Gcal)$ form in Section~\ref{sec:brm}: $\Fcal = \{Q_M^\pi: M \in \Mcal \}$, $\Gcal = \{\Tcal_{M}^\pi f: f\in\Fcal, M \in \Mcal\}$.}

\para{Structured $\Fcal$ Angle}
The above reasoning considers completely unstructured $\Fcal$. When more structured function classes (e.g., linear) are desirable,  Bellman completeness is often found satisfied in structured MDPs. (In comparison, the information-theoretic construction above does not place any restrictions on the MDP dynamics.) Example scenarios include the low-rank MDP (with a linear function class) and bisimulation abstractions (with a piecewise constant function class under the given state abstraction) \citep{chen2019information}. We present the former here as it is a very representative structural model in recent RL theory. 

\begin{example}[Low-rank MDP; \citealp{barreto2011reinforcement,barreto2014policy,jiang2017contextual}]
\label{ex:low-rank}
An MDP is a low-rank MDP with rank $\dm$, if there exists $\phi^\star: \Scal\times\Acal\to \RR^d$, $\psi^\star: \Scal\to\RR^\dm$, such that $P(s' \mid s,a) = \langle \phi^\star(s,a), \psi^\star(s')\rangle$. Also assume $R(s,a) = \phi^\star(s,a)^\top \theta_R$ for some $\theta_R \in \RR^\dm$. When $\phi^\star$ is known, the setting is called a \textit{linear MDP}  with $\phi^\star$ as its features \citep{jin2020provably}; for any $f: \RR^{\Scal\times\Acal}$ and any $\pi$, $\Tcal^\pi f$ is linear in $\phi^\star$, so  $\Fcal_{\phi^\star} = \{\langle \phi^\star, \theta \rangle: \theta\in\RR^\dm\}$, satisfies Bellman completeness.\footnote{In more detailed analyses, one often has to add norm constraints in the definition of the linear class $\Fcal_{\phi^\star}$ for concentration purposes. This adds some slight complication to Bellman completeness, since the norm of the linear coefficient may blow up after a Bellman update. This is often overcome by making appropriate norm assumptions on objects like $\phi^\star$, $\psi^\star$, and $\theta_R$; see e.g., \citep{jin2020provably}. \label{ft:norm}} When $\phi^\star$ is unknown but belongs to a feature class $\Phi$,    $\bigcup_{\phi\in\Phi} \Fcal_{\phi}$ is still Bellman complete.  
\end{example}

\para{Practical Angle} From a more practical viewpoint, the closure nature of the assumption makes it very different from realizability-type assumptions in SL: when we lack expressivity and underfit in SL, we can simply use a richer function class, hoping that the best approximation in class will be closer to the target (at least it does not hurt). However, Bellman completeness is a \textit{non-monotone} assumption, that a richer class may violate the assumption more than its subset! This, among many other challenges, makes model selection highly challenging in offline RL, which we will discuss in Section~\ref{sec:select}.  

\subsection{Bellman Residual Minimization (BRM)} \label{sec:brm}
Before we analyze FQE, we will discuss an alternative algorithm for estimating value functions. 
The divergence in Proposition~\ref{prop:deadly} can be partly attributed to the iterative nature of Fitted-Q. In comparison, in SL we often just write down a loss function and minimize it, and a consistent and ``global'' loss function is what RL is missing. Can we frame value function as loss minimization?

An immediate idea is to observe that $Q^\pi$ is the (unique) solution to $f = \Tcal^\pi f$ (Eq.\eqref{eq:bellman}), so we can find $f$ that minimizes the inconsistency of this Bellman equation. The difference, $f - \Tcal^\pi f$, known as the Bellman error (or residual), is a function of $(s,a)$. To turn that into a scalar objective, it is natural to take the expected square error on the offline data distribution $D$:
\begin{align} \label{eq:berr}
\Ecal(f; \pi) \coloneqq \EE_{D}[(f - \Tcal^\pi f)^2]. 
\end{align}
We may thus attempt to estimate $\Ecal(f; \pi)$ from data and minimize it over $f\in\Fcal$ to find $Q^\pi$, and algorithms of this kind are often referred to as Bellman residual (error) minimization (BRM) \citep{antos2008learning}.

\para{The Double-Sampling Problem} Unfortunately, the Bellman error in Eq.\eqref{eq:berr} \textit{cannot be estimated} without further assumptions \citep{farahmand2011model, sutton2018reinforcement}. The problem is that $\Tcal^\pi f$ is a conditional expectation, which is \textit{inside} square:  
$$\Ecal(f;\pi) = 
\EE_{(s,a)\sim d^D}[\left(f(s,a) - \EE_{r,s' \mid s,a}[r + \gamma f(s',\pi)]\right)^2],
$$
where $r,s' \mid s,a$ is a shorthand for $r=R(s,a), s'\sim P(\cdot \mid s,a)$. The na\"ive estimator is to simply ignore the conditional expectation  $\EE_{r,s' \mid s,a}$, which becomes $\Lhat(f;f,\pi)$ as defined below Eq.\eqref{eq:FQE}. However, this is an incorrect estimate even with infinite data (when $\Lhat (\cdot)\to \Lcal(\cdot)$), as 
$
\Lcal(f;f,\pi) \ne \Ecal(f;\pi).
$ 
More precisely,
\begin{align} \label{eq:double-sampling}
\Lcal(f;f,\pi) = \Ecal(f;\pi) + \Lcal(\Tcal^\pi f; f,\pi).
\end{align}
This is a standard bias-variance decomposition similar to regression. In SL regression when we predict real-valued label $Y$ from $X$, we also have for any $h: \Xcal \to \RR$:
\begin{align}
\EE[(Y - h(X))^2] = &~ \EE[(\EE[Y \mid X] - h(X))^2] \sts{\\
&~} + \EE[(\EE[Y \mid X] - Y)^2],
\end{align}
where the two  RHS  terms correspond to excess risk and inherent label noise, resp. However, in SL the existence of $\EE[(\EE[Y \mid X] - Y)^2]$ is not a problem even if we want to minimize the excess risk, since the inherent label noise---as its name suggests---is independent of the predictor $h$, and minimizing $\EE[(Y-h(X))^2]$ is equivalent to minimizing excess risk, even if we cannot estimate the latter.

In RL, however, the corresponding term $\Lcal(\Tcal^\pi f; f,\pi)$ \textbf{depends on the candidate function $f$ itself}, thus making the minimization of $\Lcal(f;f,\pi)$ and $\Ecal(f;\pi)$ not equivalent. This problem was identified by \citet{baird1995residual}, who proposed sampling two independent next-states from each $(s,a)$ to form an unbiased estimate of $\Ecal(f;\pi)$.\footnote{In regression, this corresponds to sampling two independent labels $Y, Y'$ from the same $X$ and estimating the excess risk as $\EE[(h(X) - Y)(h(X) - Y')]$.} However, this ``double sampling'' algorithm can only be performed in a simulator and does not apply in our setup. 

Given this problem, a fix proposed by \citet{antos2008learning} is to explicitly estimate 
$\Lcal(\Tcal^\pi f; f,\pi)$ and subtract it off $\Lcal(f;f,\pi)$. Noting that $\Lcal(\Tcal^\pi f; f,\pi)$ is the Bayes error rate of predicting $r + \gamma f(s',\pi)$ from $(s,a)$ with $\Tcal^\pi f$ as the Bayes-optimal predictor, we can write it as 
\begin{align} \label{eq:g-Tf}
\Lcal(\Tcal^\pi f; f,\pi) = \min_{g\in \RR^{\Scal\times\Acal}} \Lcal(g; f,\pi).
\end{align}
When we use a function class $\Gcal$ to model $g$, the overall estimator for $Q^\pi$ on the actual dataset $\Dcal$ becomes \citep{antos2008learning}:

\begin{mdframed}
\begin{equation}\label{eq:antos}
\begin{gathered}
\hat f^\pi = \argmin_{f\in\Fcal} \Ehat(f; \pi) \\
\textrm{where}~ \Ehat(f; \pi) \coloneqq \max_{g\in\Gcal} \, \Lhat(f;f,\pi) - \Lhat(g; f,\pi).
\end{gathered}
\end{equation}
\end{mdframed}

Clearly, $\Ehat(f;\pi)$ is only a good estimation of $\Ecal(f;\pi)$ if $\min_{g\in\Gcal}\Lcal(g;f,\pi) \approx \Lcal(\Tcal^\pi f;f,\pi)$, which requires that $\Tcal^\pi f \in \Gcal$  for all $f\in\Fcal$. Interestingly, if we use $\Fcal$ itself as $\Gcal$, this becomes $\Tcal^\pi f\in\Fcal, ~\forall f\in\Fcal$, i.e., \textit{Bellman completeness} in Assumption~\ref{asm:complete}! For presentation purposes, \textbf{we will stick with $\Gcal \equiv \Fcal$ for the rest of this article} unless otherwise noted (e.g., we will still separate $\Fcal$ and $\Gcal$ when discussing the approximation errors). 

\para{Estimation Guarantee} Under Bellman completeness, with standard concentration arguments (see e.g., \citep{xie2021bellman}), we obtain that w.p.~$\ge 1-\delta$, $\forall f\in\Fcal$, 
\begin{align} \label{eq:berr_estm}
|\Ehat(f;\pi) - \Ecal(f;\pi) | \lesssim  \frac{\Vmax^2}{n}\log\frac{|\Fcal|}{\delta},  
\end{align}
Given $Q^\pi \in \Fcal$ (a.k.a.~\textit{realizability}),\footnote{Realizability is automatically implied from completeness for $\Gcal = \Fcal$; when $\Gcal\ne\Fcal$, it needs to be assumed separately on $\Fcal$.} 
$$
\Ecal(\hat f^\pi; \pi) \le_{\epsilon} \Ehat(\hat f^\pi; \pi) \le \Ehat(Q^\pi; \pi) \le_\epsilon \Ecal(Q^\pi;\pi) = 0, 
$$ 
where $\le_\epsilon$ is $\le$ up to a small additive term that corresponds to the RHS of Eq.\eqref{eq:berr_estm}. This further  implies that 
\begin{align}
\label{eq:on-data} 
\Ecal(\hat f^\pi;\pi) \lesssim \frac{\Vmax^2}{n}\log\frac{|\Fcal| }{\delta}.
\end{align}
The result is for finite $\Fcal$; extensions to continuous $\Fcal$ with bounded $L_\infty$ covering numbers are feasible, but there are challenges with other complexity measures due to union bounding over $f$ in the TD target ($r + \gamma f(s',\pi)$), which we will discuss in Section~\ref{sec:discuss}. 

For downstream analyses, it will be convenient to introduce the notation for (distribution-)weighted $p$-norm (i.e., $\mathcal{L}^p$ norm): given a distribution $\mu$ over a space $\Xcal$,
$$
\|(\cdot)\|_{p, \mu}^p \coloneqq \EE_\mu[|\cdot|^p].
$$
This allows us to write Eq.\eqref{eq:on-data} as 
\begin{align}     \label{eq:weighted-norm}
\quad \|\hat f^\pi - \Tcal^\pi \hat f^\pi\|_{2, D} \lesssim \Vmax\sqrt{\frac{\log(|\Fcal|/\delta)}{n}}.
\end{align}
We now hope to translate this guarantee on the Bellman error of $\hat f^\pi$ to a bound on $f - Q^\pi$, and finally the estimation error of $J(\pi)$. A classical result on how Bellman error translates to value error is: $\forall f\in\RR^{\Scal\times\Acal}$,
\begin{align}\label{eq:linf}
\|f - Q^\pi\|_\infty \le \frac{ \|f - \Tcal^\pi f\|_\infty}{1-\gamma}.
\end{align}
However, just as the convergence of VI in Eq.\eqref{eq:contraction}, these $L_\infty$ results are difficult to use in the setting of learning in large state spaces: Eq.\eqref{eq:linf} requires $L_\infty$ Bellman error, but Eq.\eqref{eq:weighted-norm} only controls the much weaker weighted 2-norm. 
We will address this discrepancy in the next section and see how the notion of state (and action) coverage naturally arises when we depart from the classical $L_\infty$ analyses and consider \textit{error propagation} in Section~\ref{sec:error-prop}. 

A few remarks before we move on:

\para{Related Estimators} 
Besides BRM in Eq.\eqref{eq:antos}, there are other ways to estimate $\Ecal(f;\pi)$. One notable approach is to leverage Fenchel duality \citep{dai2018sbeed}: 
$\Ecal(f;\pi) = \min_{g\in \RR^{\Scal\times\Acal}} \EE_{D}[g(s,a)(f(s,a) - r - \gamma f(s',\pi))
- \tfrac{1}{2} g(s,a)^2)].
$ 
The equation still holds if we restrict $g\in\Gcal$ as long as $(f - \Tcal^\pi f)\in \Gcal$. Just like BRM needs $\Gcal$ to realize $\Tcal^\pi f$, these approaches require $\Gcal$ to realize Bellman error $f - \Tcal^\pi f$,\footnote{In fact it is easy to go back and forth between these two assumptions: if $\Tcal^\pi f\in \Gcal$, then $\Fcal - \Gcal \coloneqq \{f - g: f\in\Fcal, g\in\Gcal\}$ will realize all Bellman errors, and vice versa.} so the sample-complexity analyses are similar. 

\para{Computation} BRM in Eq.\eqref{eq:antos} requires solving a minimax optimization problem. When $\Fcal$ is linear, the problem has a closed-form solution that coincides with LSTDQ (see Section~\ref{sec:lstd}). When $\Fcal$ is neural net, however, the computational tractability of Eq.\eqref{eq:antos} becomes less clear; see  Section~\ref{sec:deep} for a related discussion. 

\para{Approximation Errors} So far we make exact expressivity assumptions (e.g., $Q^\pi \in \Fcal$, $\Tcal^\pi f \in \Gcal$). We can consider approximate versions, where approximation errors (e.g., $\min_{f\in\Fcal} \|f - Q^\pi\|$ for some  $\|\cdot\|$) will enter the final error bounds; how to insert them and what norm to use  are usually clear from the error propagation analyses we will see next. 
For Bellman completeness, apart from the standard additive error ($\max_{f\in\Fcal}\min_{g\in\Gcal} \|g - \Tcal^\pi f\|_{2, D}$), one can also allow a form of \textit{multiplicative} approximation in BRM analyses \citep{zanette2023realizability}. 

\subsection{Guarantee of BRM under State-Action Coverage} \label{sec:error-prop}

We are finally ready to see how value-function estimation leads to guarantees that require weaker notions of coverage compared to IS in Section~\ref{sec:IS}. We will focus on the BRM algorithm in Section~\ref{sec:brm} due to its clean analysis, and will provide a sketch for FQE afterwards. 

As alluded to at the end of Section~\ref{sec:brm}, traditional $L_\infty$ analysis is insufficient for us. Instead, we introduce a fine-grained version of Eq.\eqref{eq:linf} that is  ``distribution-aware'':

\begin{mdframed}
\begin{lemma}[\textbf{Bellman error telescoping}] \label{lem:eval_error}
For any $\pi$, and any $f\in\RR^{\Scal\times\Acal}$,
$$J_f(\pi) - J(\pi) = \frac{1}{1-\gamma}\EE_{d^\pi}[f - \Tcal^\pi f],$$
where $J_f(\pi)\coloneqq \EE_{s\sim d_0}[f(s,\pi)]$, and $d^\pi$ is the discounted state-action occupancy of $\pi$: $d^\pi = (1-\gamma)\sum_{t=0}^\infty \gamma^t d_t^\pi$, with $d_t^\pi(s,a) = \Pr_\pi[s_t = s, a_t = a]$. 
\end{lemma}
\end{mdframed}
$J_f(\pi)$ is the estimation of $J(\pi)$ if we treat $f \approx Q^\pi$, and the error is exactly equal to the average Bellman error---without any absolute value or square on each $(s,a)$---on the discounted occupancy $d^\pi$, which reflects the visitation frequency of $\pi$ over the state-action space. 

We are very close to the final guarantee: Lemma~\ref{lem:eval_error} shows that we want to control Bellman error under $d^\pi$, and Eq.\eqref{eq:weighted-norm} shows we can control the (squared) Bellman error under $d^D$. The last piece of the puzzle is a result that allows us to convert between the errors under different distributions.

\begin{mdframed}
\begin{lemma}[Error translation under coverage] \label{lem:translate}
For any function $\xi$ over space $\Xcal$, let $\mu, \nu \in \Delta(\Xcal)$ and $1\le p< \infty$, then 
$$
\|\xi\|_{p, \nu}^p \le \|\nu/\mu\|_\infty \cdot \|\xi\|_{p, \mu}^p,
$$
where $\|\nu/\mu\|_\infty \coloneqq \max_{x} \nu(x)/\mu(x)$. In this article we adopt the convention that $0/0=0$, and a non-zero value divided by $0$ is infinity.
\end{lemma}
\end{mdframed}

\para{Error Propagation under State-action Coverage} Lemma~\ref{lem:translate}  suggests a coverage assumption for BRM (Eq.\eqref{eq:antos}): 
\begin{mdframed}
\begin{assumption} \label{asm:concentrability}
Assume $\|d^\pi / d^D\|_\infty \le C_{\pi} < \infty$.
\end{assumption}
\end{mdframed}
Based on this assumption, we immediately have the first error guarantee for estimating $J(\pi)$ by  BRM : 
\begin{align}
&~ |J_{\hat f^\pi}(\pi) - J(\pi) | && \\
= &~ \left|\frac{1}{1-\gamma}\EE_{d^\pi}[\hat f^\pi - \Tcal^\pi \hat f^\pi]\right| \label{eq:telescope}
&& \text{(Lem.~\ref{lem:eval_error})} \\
\le &~ \frac{1}{1-\gamma}\|\hat f^\pi - \Tcal^\pi \hat f^\pi\|_{1, d^\pi} \label{eq:l1-relax}
&& \\
\le &~  \frac{1}{1-\gamma}\|\hat f^\pi - \Tcal^\pi \hat f^\pi\|_{2, d^\pi} \label{eq:C-S}
&& \\
\le &~ \frac{\sqrt{C_\pi}}{1-\gamma}\|\hat f^\pi - \Tcal^\pi \hat f^\pi\|_{2, d^D} \label{eq:dD-error}
&& \text{(Lem.~\ref{lem:translate})} \\
\lesssim &~ \frac{\Vmax}{1-\gamma}\sqrt{\frac{C_\pi \log(|\Fcal| /\delta)}{n}}. \label{eq:antos_error}
&& \text{(Eq.~\eqref{eq:weighted-norm})}
\end{align}

$C_\pi$ from Assumption~\ref{asm:concentrability} is also called \textit{concentrability coefficient} \citep{munos2007performance, farahmand2010error, chen2019information}, which measures coverage of $d^D$ over $d^\pi$ by the maximum density ratio. This can be significantly smaller than $(C_{\Acal})^H$ in IS (c.f.~the MDP example in Section~\ref{sec:curse}) and is consistent with the intuition given at the beginning of Section~\ref{sec:vf}, that it should suffice if we cover the state-action space instead of the trajectory space. Below we show how it can be tightened, inducing alternative notions of coverage parameters.

\para{Alternative Coverage Parameters} 
From Eq.\eqref{eq:dD-error}, it is clear that $C_\pi$ is essentially used to bound 
$
\|\hat f^\pi - \Tcal^\pi \hat f^\pi\|_{2, d^\pi}^2/\|\hat f^\pi - \Tcal^\pi \hat f^\pi\|_{2, d^D}^2, 
$ 
and can be loose when the function class $\Fcal$  has additional structures. In particular, since $\hat f^\pi \in \Fcal$, this ratio can be relaxed to an a priori (i.e., data-independent) quantity, which is a drop-in improvement for $C_\pi$ in the above analysis:
\begin{align} \label{eq:sq-to-sq}
\Csq_\pi \coloneqq \max_{f\in\Fcal} \frac{\|f - \Tcal^\pi f\|_{2, d^\pi}^2}{\|f - \Tcal^\pi f\|_{2, d^D}^2}. 
\end{align}
(Recall that $\|\cdot\|_{2, d^D}^2 = \EE_{d^D}[(\cdot)^2]$.) 
When $\Fcal$ is   the linear class induced by a feature map $\phi\in\RR^{\dm}$, i.e., $\Fcal \subseteq \{\phi(s,a)^\top \theta\}$, Bellman completeness implies that $\Tcal^\pi f \in \Fcal$ and thus $f - \Tcal^\pi f$ are also linear, in which case Eq.\eqref{eq:sq-to-sq} has a very interpretable upper bound:
\begin{align} \label{eq:linear-sq}
\qquad \Csq_\pi \le \max_{u\in \RR^{\dm}} \frac{u^\top \Sigma_{\pi} u}{u^\top \Sigma_{D} u} = \sigma_{\max}( \Sigma_{\pi}^{1/2} \Sigma_D^{-1} \Sigma_{\pi}^{1/2}),
\end{align}
where $\Sigma_\pi = \EE_{d^\pi}[\phi\phi^\top]$ and $\Sigma_D$ is defined similarly for $d^D$, and $\sigma_{\max}$ denotes the largest eigenvalue. This coverage parameter only requires $(s,a)\sim d^D$ to hit all \textit{feature directions} activated by $d^\pi$, and can be bounded even if $\|d^\pi/d^D\|_\infty$ is infinity. 

In fact, we can further tighten the coverage parameter by directly translating Eq.\eqref{eq:telescope} to on-data error $\|f - \Tcal^\pi f\|_{2, d^D}$ (note that the square on the numerator is now \textit{outside} the expectation) \citep{duan2020minimax, song2022hybrid}:
\begin{align} \label{eq:avg-to-sq}
\Cavg_\pi \coloneqq \max_{f\in\Fcal} \frac{(\EE_{d^\pi}[f - \Tcal^\pi f])^2}{\EE_{d^D}[(f - \Tcal^\pi f)^2]}.
\end{align}
In the same linear setting as above, we have \citep{duan2020minimax, yin2020asymptotically, zanette2021provable}: 
\begin{align} \label{eq:linear-avg}
\Cavg_\pi \le \EE_{d^\pi}[\phi]^\top \Sigma_D^{-1} \EE_{d^\pi}[\phi].
\end{align}
This is a very notable improvement over Eq.\eqref{eq:linear-sq}, as we now only need to cover a single direction in $\RR^{\dm}$, the \textit{mean} feature under $d^\pi$! This shows that the Cauchy-Schwartz step in Eq.\eqref{eq:C-S} can be surprisingly loose sometimes. That said, not all settings and methods permit the tight definition of coverage in Eq.\eqref{eq:avg-to-sq}, and sometimes we need to resort to Eq.\eqref{eq:sq-to-sq} or similar quantities (Section~\ref{sec:pevi}). For example, the $\hat f^\pi$ learned by BRM under $\Cavg_\pi$ coverage only guarantees accurate $J(\pi)$ estimation; if we want stronger guarantees such as bounded $\EE_{\nu}[(f - Q^\pi)^2]$ for some distribution $\nu$ (i.e., $f$ has the correct ``shape'' on $\nu$),
we will need $d^D$ to cover $d_\nu^\pi$ (i.e., occupancy induced by $\nu$ as the initial distribution) in the $\Csq$ sense. 

Final comments about coverage:
\begin{enumerate}[leftmargin=*]
\item When we relax $\{f - \Tcal^\pi f : f\in\Fcal \}$ to $\RR^{\Scal\times\Acal}$ in $\Csq_\pi$ and $\Cavg_\pi$, we recover a definition that is based on raw densities, but slightly different from and always upper bounded by $C_\pi$: $\EE_{d^D}[(d^\pi/d^D)^2]$ \citep{xie2020q}. Therefore, apart from leveraging the structure of $\Fcal$, $\Csq_\pi$ and $\Cavg_\pi$ also contain  an implicit improvement of using second moment instead of infinity-norm to measure the size of the density ratio function. 
\item Another common assumption is full-rank $\Sigma_D$ and $\sigma_{\min}(D)$ bounded away from $0$. Combined with an upper bound on $\|\phi\|_2$, we can immediately bound $\Csq_\pi$ and $\Cavg_\pi$ using $1/\sigma_{\min}(\Sigma_D)$, \textit{regardless of $\pi$}. Compared to $C_\pi$, this assumption is weaker in that it leverages the linearity of $\{f - \Tcal^\pi f\}$, but is also stronger in ignoring the properties of $d^\pi$. In fact, when $\dm = \Scal\times\Acal$  and $\phi$ is the $(s,a)$-indicator feature, $\Sigma_D = \textrm{diag}(\{d^D(s,a)\}_{(s,a)\in\Scal\times\Acal})$, so $\sigma_{\min}(\Sigma_D) = \min_{s,a} d^D(s,a)$, a type of parameter often referred to as \textit{reachability}. Note that this quantity \textit{necessarily} scales with $1/|\Scal\times\Acal|$, whereas $C_\pi$ does not. 
\end{enumerate}

\para{FQE Analysis}
For FQE in Section~\ref{sec:fqe}, we can establish an estimation error bound on $\|f_k - \Tcal f_{k-1}\|_{2,D}$ under Bellman completeness that is similar to Eq.\eqref{eq:weighted-norm}. The difference is that $f_k$ is not Bellman-consistent with itself, but with the function in previous iteration, $f_{k-1}$. This makes the error propagation analysis (Section~\ref{sec:error-prop}) slightly more involved for FQE, but the overall idea is similar to BRM and we give a sketch here. 

The key observation is that running FQE for $K$ iterations can be viewed as learning the non-stationary value function for a truncated finite-horizon MDP, where the return is defined as $J_K(\pi)\coloneqq \EE_{\pi}[\sum_{t=0}^{K-1} \gamma^t r_t]$.\footnote{Here $f_0 \equiv 0$ is assumed, and non-zero $f_0$ can be added to the objective and does not affect the analysis.} $J_K(\pi)$ well-approximates the infinite-horizon objective $J(\pi)$ up to a residual term controlled by $\gamma^K$, and can be made arbitrarily small by choosing large $K$. Value function for this objective is time-dependent: $Q_k^\pi(s,a) \coloneqq \sts{\textstyle} \EE_{\pi}[\sum_{t=0}^{k-1} \gamma^t r_t \,|\, s_0 = s, a_0 = a]$, and $J_{Q_K^\pi}(\pi) = J_K(\pi)$. The non-stationary Q-function also satisfies Bellman equation: $Q_k^\pi = \Tcal^\pi Q_{k-1}^\pi$, with $Q_0^\pi \equiv 0$. FQE's output $f_K, \ldots, f_1$ are approximations of $Q_K^\pi, \ldots, Q_1^\pi$, and small $\|f_k - \Tcal^\pi f_{k-1}\|_{2, D}$ thus implies the Bellman consistency of $\{f_K, \ldots, f_1\}$ as a single time-dependent function in the finite-horizon problem. Based on this understanding, we can write down the finite-horizon variant of Lemma~\ref{lem:eval_error}: 
\begin{lemma}\label{eq:telescope-non-stat}
For any non-stationary policy  $\pi_{K:1}$ and function $f_{K:1}$, we have $J_{f_K}(\pi_K) - J_K(\pi_{K:1})=  $
\begin{align}
\sum_{t=0}^{K-1} \gamma^{t} \EE_{d_t^{\pi_{K:1}}}[f_{K-t} - \Tcal^{\pi_{K-t-1}} f_{K-t-1}].
\end{align}
\end{lemma}
The form applies to a $K$-step non-stationary policy $\pi_{K:1}$, which takes $a_t$ according to $\pi_{K-t}(\cdot \mid s)$; for now we only need $\pi_K=\cdots=\pi_1 = \pi$, and the general form will be of use later. The rest of the analysis follows similarly as Eqs.\eqref{eq:l1-relax}--\eqref{eq:antos_error}, except that we now need $d^D$ to cover $d_t^\pi$ separately for each $t$ instead of covering $d^\pi$ as a whole \citep{munos2007performance, farahmand2010error, xie2020q}.

\begin{figure*}[t]
\centering
\includegraphics[width=\sts{.75}\arx{.46}\columnwidth]{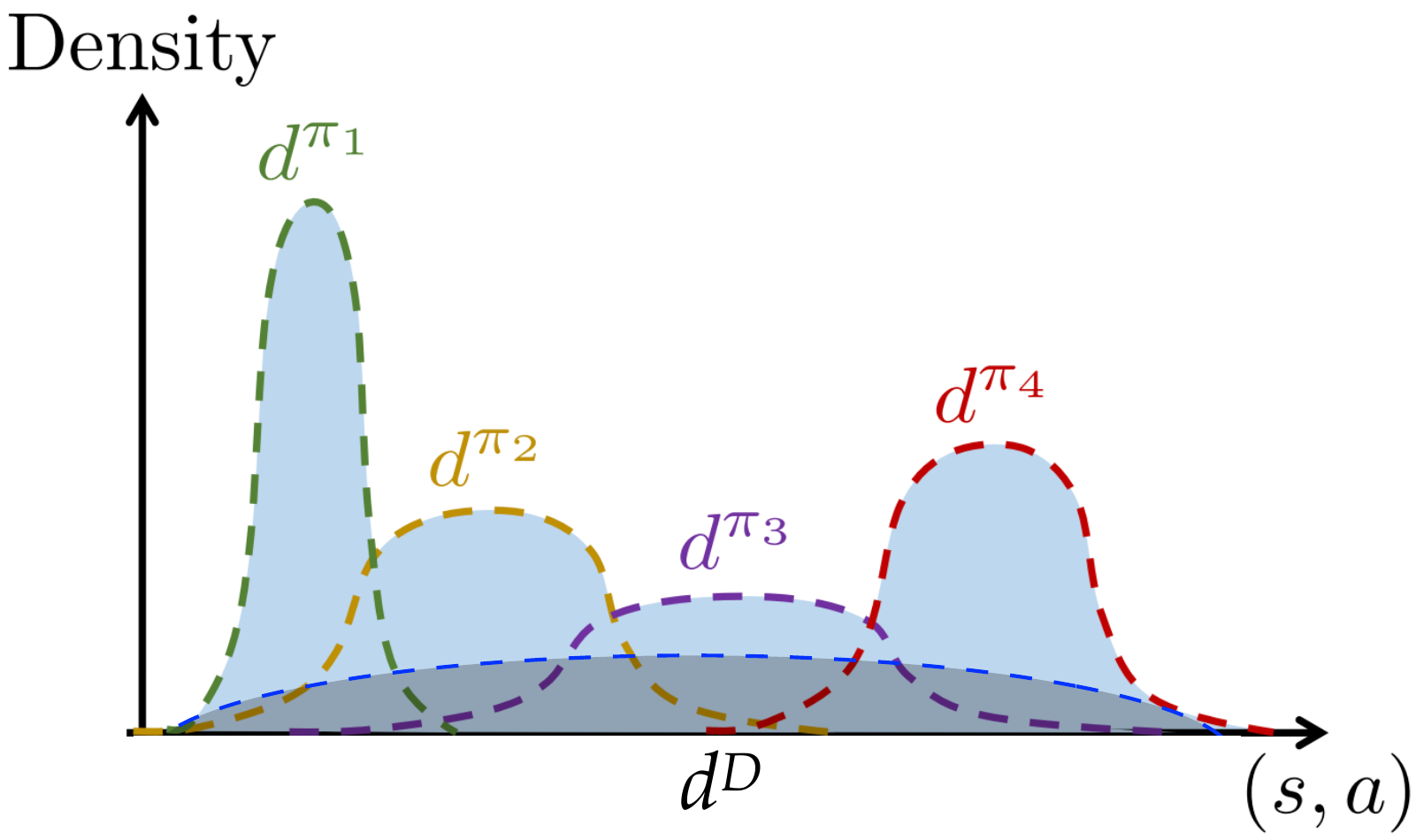} \hspace{3em}
\includegraphics[width=\sts{.75}\arx{.46}\columnwidth]{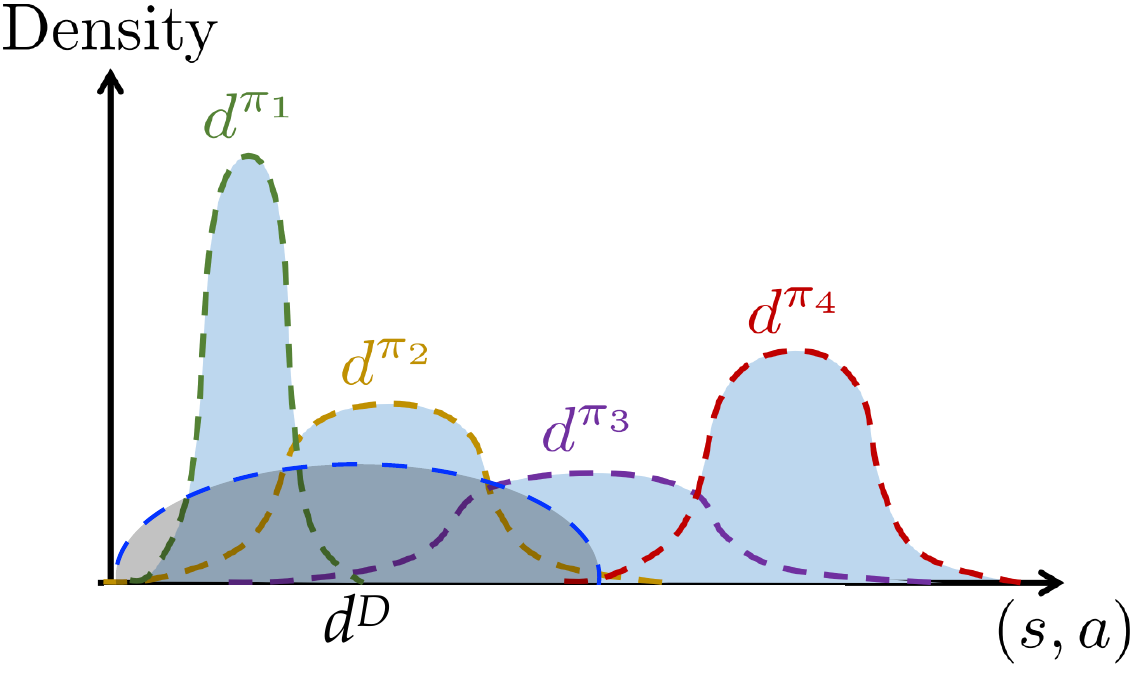}
\caption{Figurative illustration of different coverage assumptions, adapted from \citet{xie2022role}. \textbf{Left:} All-policy coverage. \textbf{Right:} Data only covers $\pi_1$ and $\pi_2$, and pessimistic algorithms in Section~\ref{sec:pess} can compete with the best among them. \label{fig:coverage}}
\end{figure*}

\subsection{Policy Optimization} \label{sec:neutral_po}
So far we have focused on OPE, which is crucial for holdout validation and testing (we will discuss more in Section~\ref{sec:select}). For training, however, we will need to perform policy optimization. An immediate reduction is 
\begin{align} \label{eq:opl_neutral_info}
    \argmax_{\pi \in \Pi}J_{\hat f^\pi}(\pi),
\end{align}
(see the definition of $J_f(\pi)$ in Lemma~\ref{lem:eval_error}) where $\hat f^\pi$ is estimated from Eq.\eqref{eq:antos} or other methods. Then, we can upper bound the prediction error from Eq.\eqref{eq:antos_error} uniformly over $\pi\in\Pi$, which immediately translates to the following policy optimization guarantee:

\begin{theorem} \label{thm:neutral}
Let $\hat \pi$ be the output of Eq.\eqref{eq:opl_neutral_info} with $\Gcal=\Fcal$. Suppose Assumptions~\ref{asm:complete} and \ref{asm:concentrability} hold for all $\pi \in \Pi$, then w.p.~$\ge 1-\delta$, for any $\picomp\in \Pi$,
$$
J(\picomp) - J(\hat\pi) \lesssim \frac{\Vmax}{1-\gamma}\sqrt{\frac{(\max_{\pi\in\Pi} C_\pi) \log(|\Fcal| |\Pi|/\delta)}{n}}.
$$
\end{theorem}
Here $\picomp$ is any policy we may wish to compete with and can be set to $\argmax_{\pi\in\Pi} J(\pi)$; we present in this form so that it can be  easily compared with the guarantee of improved algorithms in Section~\ref{sec:pess}. 
The additional $|\Pi|$ comes from union bounding the event of accurate $J(\pi)$ estimation across all $\pi\in\Pi$, and can be extended to continuous policy classes with appropriate notions of covering numbers \citep{antos2008learning,xie2021bellman,zanette2021provable,cheng2022adversarially}. $C_\pi$ in the bound can also be replaced with tighter definitions of coverage, such as $\Cavg_\pi$ in Eq.\eqref{eq:avg-to-sq}.

\para{All-policy Coverage} The $\max_{\pi} C_\pi$ term requires that the data distribution provides sufficient coverage over \textit{all} policies $\pi \in \Pi$ (see Figure~\ref{fig:coverage}), which puts a heavy burden on $d^D$ that it needs to be very exploratory. Worse still, there are settings where $\max_{\pi\in\Pi} C_\pi$ \textit{will be exponentially large (e.g., linear in $|\Pi|$ or exponentially in $H$) even for the best-case $d^D$}. A simple example is where the MDP is a deterministic complete tree and each path is a deterministic policy, and $\max_{\pi\in\Pi} C_\pi \ge |\Pi| = |\Acal|^H$ for any $d^D$ \citep{chen2019information}. 

That said,   certain structures in the dynamics may allow for small $\max_{\pi\in\Pi} C_\pi$ even when the state space is large. An example is   the low-rank MDP in Example~\ref{ex:low-rank}  \citep{chen2019information}: given any policy class $\Pi$, there always exists $d^D$ such that $\max_{\pi\in\Pi} C_\pi \le |\Acal| \dm$. The construction is based on the observation that all $d^\pi(s)$ in a low-rank MDP is linear in $\psi(\cdot)$, and a mixture of the barycentric spanner of $\{d^\pi: \pi\in\Pi\}$ will guarantee state coverage of $\dm$, and  the additional $|\Acal|$ comes from taking uniform actions.

\para{Computationally-efficient Algorithms}   Eq.\eqref{eq:opl_neutral_info} is an information-theoretic objective: implementing it in a literal manner would require enumerating over the policy class $\Pi$, which is typically a combinatorial object and thus leads to intractable computation. 

There are computationally more feasible algorithms that give comparable guarantees under similar or somewhat stronger assumptions. They are typically based on dynamic-programming (DP) algorithms such as value iteration and policy iteration. 

\para{{\it \textbf{Fitted-Q Iteration (FQI)}}} Besides the policy-specific Bellman equation (which we used to approximate $Q^\pi$), there is also the Bellman optimality equation, $Q^\star = \Tcal Q^\star$, where $\forall f\in\RR^{\Scal\times\Acal}$, 
\begin{align}
(\Tcal f)(s,a) \coloneqq 
R(s,a) + \gamma \EE_{s'\sim P(\cdot \mid s,a)}[\max_{a'\in\Acal} Q^\star(s',a')].
\end{align}
The fixed point of $\Tcal$, namely $Q^\star$, induces a greedy policy $\pi_{Q^\star}(s) \coloneqq \argmax_{a\in\Acal} Q^\star(s,a)$, which achieves optimal return in this MDP for all starting states simultaneously and is also denoted as $\pi^\star$.

Therefore, we may approximate the fixed point of $\Tcal$ in exactly the same way as we do to $\Tcal^\pi$, i.e., either in an iterative manner as in Eq.\eqref{eq:FQE}, or by solving a minimax optimization problem as in Eq.\eqref{eq:antos}. 
After obtaining $\hat f \in \Fcal$, we finally output its greedy policy $\pi_{\hat f}(s) \coloneqq \argmax_{a\in\Acal} \hat f(s,a)$. This way, the algorithm does not maintain a standalone policy class, and instead induces it from the value function class $\Fcal$ as $\Pi = \{\pi_f: f\in\Fcal\}$. 
The iterative algorithm, known as FQI \citep{ernst2005tree}, is more computationally tractable and can be viewed as the theoretical prototype of empirically popular algorithms such as Q-learning (or DQN in deep RL \citep{mnih2015human}).  

Despite FQI being more computationally friendly, we sketch the analysis for the minimax variant below due to its simplicity; the FQI analysis is similar in spirit \citep{munos2007performance, munos2008finite, chen2019information}. 
For the minimax algorithm, an analysis similar to the one for Eq.\eqref{eq:antos} can bound $\EE_{d^D}[(\hat f - \Tcal \hat f)^2]$, under a variant of the Bellman-completeness assumption
$$
\Tcal f \in \Fcal, \forall f \in \Fcal.
$$
The following lemma sheds light on its error propagation (see Lemma 3.1 of \citet{jin2020pessimism} for the finite-horizon version):

\begin{mdframed}
\begin{lemma} \textbf{\emph{(Generalized Performance-Difference}}   \textbf{\emph{Lemma)}} \label{lem:opl_error_prop}
For any $f\in \RR^{\Scal\times\Acal}$, $\pi,\pi': \Scal\to\Delta(\Acal)$, \vspace{-1em}
\begin{align}
J(\pi') - J(\pi) =&~  \frac{1}{1-\gamma}\big( \underbrace{\EE_{s \sim d^{\pi'}}[f(s,\pi') - f(s,\pi)]}_{\textrm{(i): Advantage}} \sts{\\ 
&~} + \underbrace{\EE_{d^{\pi'}}[\Tcal^\pi f - f]}_{\textrm{(ii): BE under $d^{\pi'}$}}  +  \underbrace{\EE_{d^{\pi}}[f - \Tcal^\pi f]}_{\textrm{(iii): BE under $d^{\pi}$}} \big).
\end{align}
\end{lemma}
\end{mdframed}
This is a key lemma for analyzing policy-optimization algorithms, and the analyses in the subsequent text will essentially control each of the terms on the RHS by either algorithm designs or additional assumptions. 
We will use this result again in its general form later, but for now we only need its corollary:\footnote{Another notable special case of this lemma is when $f = Q^\pi$.  The Bellman error (BE) terms vanish and we recover the performance difference (PD) lemma \citep{kakade2002approximately}. \label{ft:PD}} when we choose $\pi' = \pi^\star$ and $\pi = \pi_f$, the advantage term (i) is non-positive and $\Tcal^\pi f = \Tcal f$, leading to the bound that
$$
J(\pi^\star) - J(\pi_f) \le \frac{1}{1-\gamma}\big( \EE_{d^{\pi^\star}}[\Tcal f - f]  + \EE_{d^{\pi_f}}[f - \Tcal f] \big).
$$
This result is the counterpart of Lemma~\ref{lem:eval_error} but for learning $Q^\star$, which tells us that $J(\pi^\star) - J(\pi_{\hat f})$ will be small if (1) $\EE_{d^D}[(\hat f - \Tcal \hat f)^2]$ is small (which we already have from above), and (2) $d^D$ covers both $d^{\pi^\star}$ and $d^{\pi_{\hat f}}$ (in the sense of bounded $C_{(\cdot)}$ or its refined versions). The former is very reasonable and inevitable: to learn the near optimal policy, it is natural that our data should contain information about it. The latter requires $d^D$ to cover the \textit{learned policy} itself, which is a random quantity we have no direct control over. To come up with an assumption that is independent of the data randomness, we may relax to all possible policies of the form $\pi_{f}$ and pay $\max_{f\in\Fcal} C_{\pi_f}$ in the sample complexity guarantee, which coincides with the ``all-policy coverage'' term $\max_{\pi\in\Pi} C_\pi$ in Theorem~\ref{thm:neutral}. 

\para{\textbf{\textit{Fitted Policy Iteration (FPI)}}} Besides value iteration, another fundamental planning algorithm for MDPs is called \textit{policy iteration}: $\pi_{k+1} \gets \pi_{Q^{\pi_k}}$, i.e., producing the next policy to be greedy w.r.t.~the previous policy's Q-function, and FPI is the algorithm where $Q^{\pi_k}$ is approximated by algorithms introduced earlier in function class $\Fcal$. Similar to the FQI case, FPI also has its policy class induced by $\Fcal$: $\{\pi_f: f\in\Fcal\}$. The guarantee  for FPI is similar to FQI, and we will also see the analysis of a variant of FPI in Section~\ref{sec:pspi}.

\section{Pessimistic Policy Optimization}
The policy optimization guarantees we have seen so far all require all-policy coverage and incur dependence on $\max_{\pi\in\Pi} C_\pi$ ($C_\pi$ can be replaced by its refined versions). This requires the dataset to be exploratory which may not hold in practice, and sometimes even implies restrictions on the underlying MDP dynamics (Section~\ref{sec:neutral_po}). Therefore, a natural question, which has become a central consideration for offline RL research, is whether we can develop algorithms and guarantees that work for an arbitrary offline dataset (see Figure~\ref{fig:coverage} right). 

Below we will show that by using \textit{pessimistic} algorithms, we can relax all-policy coverage to \textit{single-policy} coverage: we can compete with any policy that receives good coverage by the offline data. 

\subsection{Pessimism in Face of Uncertainty} \label{sec:pess}
Consider a simple multi-armed bandit problem, which can be viewed as an MDP with only one state (so transition always returns to the state itself) and each action (or arm) yields a stochastic reward. For each $a\in\Acal$ we denote the true mean reward as $R(a)$. The candidate policies correspond to choosing different arms deterministically.  On this problem instance, the various policy-optimization algorithms in the previous section all reduce to the following simple procedure: 
\begin{enumerate}
\item Estimate the mean reward for each arm ($\widehat R(a)$).
\item Output the arm with the highest estimated mean.
\end{enumerate}

\begin{figure}[t]
\centering
\includegraphics[height=12em]{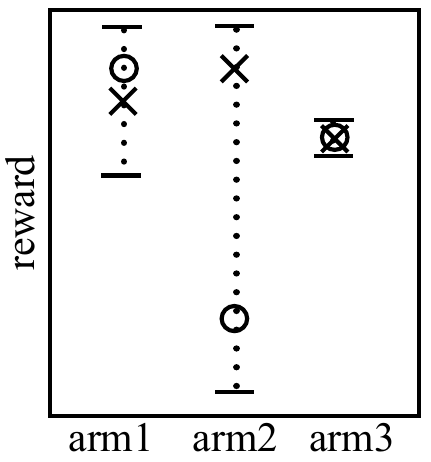} \qquad 
\includegraphics[height=12em]{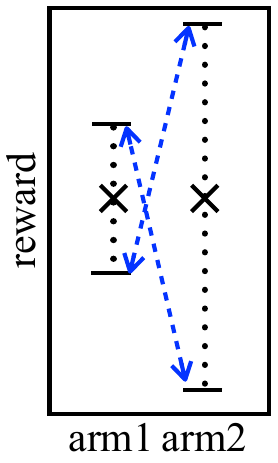}
\caption{Uncertainty in Multi-armed bandits (MABs). ``O'' is true mean and ``X'' is point estimate. \textbf{Left:} Example where greedy w.r.t.~point estimate chooses arm2 and suffers large loss. Instead, pessimism chooses arm3 with suboptimality bounded by the uncertainty of the best arm (arm1). \textbf{Right:} Example where return optimization chooses a different pessimistic policy (arm1) than regret minimization. Both arms have the same regret (height of double-headed arrows), and randomizing between them only incurs half of the regret. \label{fig:mab}}
\end{figure}

All-policy coverage requires that each arm is sampled a sufficient number of times in the offline data, so that the reward estimation is accurate for all arms. When it fails to hold, we may choose an arm with low reward if it has received very few samples due to random fluctuation, even when we have abundant samples for another high-reward arm; see Figure~\ref{fig:mab}.

The key to addressing this issue is \textit{uncertainty quantification}: instead of just looking at the point estimates, we should also consider \textit{confidence intervals} (CIs) for the estimate. In the bandit case, we may form $[R^-(a), R^+(a)]$ for each arm by concentration inequalities, and guarantee that the true mean $\mu(a) \in [R^-(a), R^+(a)]$. To reliably compete with a high-sample, high-reward arm, we may choose the arm with the highest \textit{lower-confidence bound} (LCB), $R^-(a)$. The guarantee of algorithms based on point estimates pays for the worst-case estimation error across all arms; in comparison, the LCB algorithm only pays for the uncertainty on the optimal arm: let $\hat a = \argmax_{a\in\Acal} R^-(a)$ and $a^\star = \argmax_{a\in\Acal} R(a)$,
\begin{align} \label{eq:lcb}
&~ R(a^\star) - R(\hat a)  \\
\le &~ R^-(a^\star) - R^-(\hat a) +  R(a^\star) - R^-(a^\star) \\
\le &~ R(a^\star) - R^-(a^\star).
\end{align}
This a general lesson that applies broadly: \textbf{when we optimize lower confidence bounds, the guarantee only pays for how much we under-estimate the optimal policy.} 
In fact, even when the optimal arm has a large CI, we may still achieve nontrivial guarantees by replacing $a^\star$ in Eq.\eqref{eq:lcb} with any other comparator arm that has good (albeit suboptimal) mean reward and a small CI. 

This is an example of the principle of \textit{pessimism in face of uncertainty}, which is the opposite to the well-known optimistic principle in online RL: optimism encourages exploration (i.e., leaving the current data distribution), whereas pessimism encourages exploitation (i.e., staying within the current data distribution). 

\subsection{Pessimism in MDPs: A Na\"ive Attempt}
The next question is how to instantiate this principle for MDPs. One natural idea is to consider uncertainty quantification for $J(\pi)$: if we know $J(\pi)$ falls into an interval $[J^-(\pi), J^+(\pi)]$ computed from data (e.g., the CI for each arm in the bandit example), we may choose the policy that maximizes the lower confidence bound: $\argmax_{\pi\in\Pi} J^{-}(\pi)$. One such interval is directly given by the guarantee of BRM, i.e., 
$$
J_{\textrm{cov}}^{\pm}(\pi) \coloneqq J_{\hat f^\pi}(\pi) \pm \textrm{eb}(C_\pi),
$$
where $\textrm{eb}(C_\pi)=O\left(\frac{\Vmax}{1-\gamma}\sqrt{\frac{C_\pi \log(|\Fcal| |\Pi|/\delta)}{n}}\right)$ is the error bound from Eq.\eqref{eq:antos_error} after union bounding over $\Pi$. Recall that this is a valid bound whenever we have Bellman completeness (Assumption~\ref{asm:complete}), $\Tcal^\pi f\in\Fcal,\, \forall f\in\Fcal, \pi\in\Pi$. 
By the same logic as in Eq.\eqref{eq:lcb}, the policy $\hat \pi$ that optimizes $J_{\textrm{cov}}^-(\pi)$ can directly compete with any policy $\picomp \in \Pi$ (such as $\argmax_{\pi\in\Pi}J(\pi)$) under the desired single-policy coverage condition:
\begin{align}
 J(\picomp) - J(\hat \pi) 
\le &~ J_{\textrm{cov}}^-(\picomp) + \textrm{eb}(C_{\picomp}) - J_{\textrm{cov}}^-(\hat \pi) \\
\le &~ \textrm{eb}(C_{\picomp}). \label{eq:single-policy}
\end{align}
The last line follows from the fact that $\hat \pi$ maximizes $J_{\textrm{cov}}^-(\cdot)$. Compared to Theorem~\ref{thm:neutral}, the bound successfully replaced $\max_{\pi\in\Pi} C_{\pi}$ with $C_{\picomp}$. 

The problem with the approach is that the error bound $\textrm{eb}(C_\pi)$ depends on $C_\pi$. $C_\pi$ is known in the bandit example (which is the inverse of the probability of choosing an arm in the offline data), but is generally inaccessible as it involves the discounted occupancy $d^\pi$ that depends on MDP dynamics (see Lemma~\ref{lem:eval_error}). Next we show how to obtain the same guarantee  without knowing $C_\pi$. 

\subsection{Version-space pessimism}  \label{sec:vs-pessimism}
Recall from Eq.\eqref{eq:antos} that we minimized the following loss over $f\in\Fcal$ to obtain $\hat f^\pi$:
\begin{align}
\Ehat(f; \pi)\coloneqq \max_{g\in\Fcal} \, \Lhat(f;f,\pi) - \Lhat(g; f,\pi),
\end{align}
as $\Ehat(f;\pi) \approx \Ecal(f;\pi) = \|f - \Tcal^\pi f\|_{2, D}^2$ is small for $f = Q^\pi$. This immediately implies that 
w.p.~$\ge 1-\delta$, $\forall \pi\in\Pi$, 
\begin{mdframed} \vspace{-1em}
\begin{align} \label{eq:vs}
Q^\pi \in \VS{\pi}\coloneqq \{f\in \Fcal: \Ehat(f; \pi) \le \epsilon_0\}, 
\end{align}
\end{mdframed}
where $\epsilon_0$ is the RHS of Eq.\eqref{eq:berr_estm} (plus union-bounding over $\Pi$) to make sure that $Q^\pi$ is not excluded for all $\pi\in\Pi$. We call $\VS{\pi}$ the \textit{version space} for $Q^\pi$. This allows us to come up with a lower confidence bound for $J(\pi)$: 
\begin{align} \label{eq:Jvs-}
& J^-_{\textrm{VS}}(\pi) \coloneqq \min_{f\in\VS{\pi}} J_f(\pi) \le J_{Q^\pi}(\pi) = J(\pi). 
\end{align}
We then optimize this objective over $\pi\in\Pi$.  
Similar to the analysis of $\argmax_{\pi\in\Pi}J_{\textrm{cov}}^-(\pi)$, we will pay for an upper bound on the difference between $J(\picomp)$ and $J^-_{\textrm{VS}}(\picomp)$. 
To bound $J(\picomp) - J^-_{\textrm{VS}}(\picomp)$, we combine the definition of $\VS{\pi}$ with Eq.\eqref{eq:berr_estm} and immediately have that
$$
\Ecal(f;\pi) \le 2\epsilon_0, ~ \forall f\in \VS{\pi}.
$$
This also holds for $f_{\min}^\pi\coloneqq \argmin_{f\in\VS{\pi}}J_f(\pi)$ since $f_{\min}^\pi \in \VS{\pi}$. Using the same telescoping and error translation analysis as Eqs.\eqref{eq:telescope}--\eqref{eq:antos_error}, we have
$$
J(\picomp) - J^-_{\textrm{VS}}(\picomp) \le \frac{\sqrt{C_\pi}}{1-\gamma} \sqrt{2\epsilon_0}. 
$$

Plugging in the expression for $\epsilon_0$, we have the more formal guarantee:
\begin{theorem} \label{thm:pessimism}
Fix any $\picomp\in\Pi$. Assume (1) Assumptions~\ref{asm:concentrability} and \ref{asm:complete} for $\pi=\picomp$, and (2) $Q^\pi \in \Fcal$ for all $\pi\in\Pi$. Then, the policy $\hat\pi$ that maximizes $J_{\textrm{VS}}^-(\pi)$ over $\pi\in\Pi$ enjoys the following guarantee: w.p.~$\ge 1-\delta$, 
$$
J(\picomp) - J(\pi) \lesssim \frac{\Vmax}{1-\gamma}\sqrt{\frac{C_{\picomp} \log(|\Fcal| |\Pi|/\delta)}{n}}.
$$
\end{theorem}
Similar to before, $C_{\picomp}$ can be replaced with its refined versions such as $\Cavg_{\picomp}$. Compared to Theorem~\ref{thm:neutral}, we have relaxed all-policy coverage ($\max_\pi C_\pi$) to single-policy coverage ($C_{\picomp}$); in fact, the guarantee can be non-vacuous even when the optimal policy is not covered, as we can simply choose any other comparator policy $\picomp$ that is covered by data. 

Theorem~\ref{thm:pessimism} has also relaxed the expressivity assumptions: it only requires Bellman-completeness for $\picomp$, as that guarantees tightness of the lower bound $J_{\textrm{VS}}^-(\picomp)$ which we pay in the guarantee. On the other hand, for any other policy $\pi$, all we need is the \textit{validity} of $J_{\textrm{VS}}^-(\pi)$, i.e., that it is actually a lower bound of $J(\pi)$. As long as $Q^\pi \in \Fcal$, we have $\Tcal^\pi Q^\pi = Q^\pi \in \Fcal$, which implies that $\Ecal(Q^\pi; \pi) \approx   0$ even when $\Fcal$ does not satisfy Bellman completeness for $\pi$. As a result, $Q^\pi$ will never be eliminated from $\Fcal_{\epsilon_0}^\pi$, which guarantees the validity of $J_{\textrm{VS}}^-(\pi)$. 

\subsection{An Oracle-efficient Algorithm: PSPI} \label{sec:pspi}
Despite the significant improvement in coverage, the algorithm is information-theoretic (in the same way as Eq.\eqref{eq:opl_neutral_info}), and here we introduce a more computationally friendly version. In RL with function approximation, computational efficiency is often stated in the form of \textit{oracle efficiency}, i.e., the computation is efficient if we are given blackbox oracles for certain optimization subroutines that can be (1) reasonably approximated in practice, and (2)  efficiently implemented  (without further assumptions) in special cases such as tabular and linear classes. 

The FQE and FQI algorithms are examples where the oracle assumption is very straightforward: least-square regression oracles over $\Fcal$. For pessimism, however, we require the oracle of computing the worst-case $Q$-function for any given policy $\pi$:
\begin{align} \label{eq:fmin}
f_{\min}^\pi \coloneqq \argmin_{f\in\Fcal_{\epsilon_0}^\pi} J_f(\pi). 
\end{align}
Recall that $f\in\Fcal_{\epsilon_0}^\pi \Longleftrightarrow \Ehat(f;\pi) \le \epsilon_0$, so this is essentially a constrained optimization problem. We may turn it into a ``Lagrangian''-like version\footnote{This is not strictly speaking a Lagrangian because $\lambda$ is fixed and not being optimized. This reflects a more general situation in RL theory, that computationally efficient algorithms are often \textit{not} designed by (1) taking an information-theoretic algorithm, and (2) implementing or approximating it with bounded computation. Rather, the algorithm will take inspiration from the info-theoretic algorithm but have distinct statistical behaviors, which require new analyses.}  $\min_{f\in\Fcal} J_f(\pi) + \lambda \Ehat(f;\pi)$, which results in a minimax optimization problem due to the $\max_g$ in the estimation of $\Ehat(f;\pi)$. \citep{xie2021bellman} further showed that this program is fully efficient when $\Fcal$ is linear in feature map $\phi(s,a)$, as $\lambda \Ehat(f;\pi)$ has a close-form expression that is quadratic in $\phi$ with a PSD Hessian (this bears a close connection to LSTDQ discussed in Section~\ref{sec:lstd}). Given that $J_f(\pi)$ is also linear in $f$ (thus in $\phi$), the overall problem is an instance of convex quadratic programming.\footnote{Similar to Footnote~\ref{ft:norm}, here we also need norm constraints on the linear coefficients to control the boundedness of functions in $\Fcal$, but that only adds a few more convex constraints to the convex program.}

\para{PSPI} Given such an oracle,  \citep{xie2021bellman} propose the following oracle-efficient algorithm that achieves guarantees under single-policy coverage: 

\begin{mdframed}
\begin{enumerate}[leftmargin=*]
\item Initialize $\pi_1$ as uniformly random over $\Acal$.
\item \textbf{For} $k=1, 2, \ldots, K$, 
\begin{enumerate}
    \item Compute $f_k\coloneqq f_{\min}^{\pi_k}$ using the oracle.
    \item $\pi_{k+1}(a \mid s) \propto \pi_k(a \mid s)\exp(\eta f_k(s,a))$.
\end{enumerate}
\item Output uniform mixture of $\pi_1, \ldots \pi_K$, $\textrm{Unif}[\pi_{1:K}]$. 
\end{enumerate}
\end{mdframed}
The ``uniform mixture'' is \textit{trajectory-level} mixture, i.e., when executing the policy, we will randomly sample an integer $i$ uniformly between $1$ to $K$, and roll out the trajectory with policy $\pi_i$. The number $i$ only gets resampled when another trajectory is generated. In general, such trajectory-level mixture creates non-Markov and history-dependent policies, even if the ``base policies'' ($\pi_1, \ldots, \pi_K$) themselves are Markov, so this can be viewed as an instance of \textit{improper learning}, which is common when no-regret algorithms are incorporated in RL \citep{ross2011reduction}. As a direct consequence, $J(\textrm{Unif}[\pi_{1:K}]) = 1/K \sum_{k=1}^K J(\pi)$. 

Computationally, apart from calling the oracle in Eq.\eqref{eq:fmin}, we also need to execute the policy update step 2b). Here we no longer use a standalone policy class, but have the policy class induced from the value function class $\Fcal$ similar to FQI and FPI in Section~\ref{sec:neutral_po}. The difference is that instead of hard argmax (i.e., greedy) over $f\in\Fcal$ we are now taking softmax policies over mixture of $f\in\Fcal$, with the following implicit policy class:
$$
\Pi = \arx{\left}\{ \pi(\cdot \mid s) \propto \exp\arx{\left}(\eta \textstyle \sum_{i=1}^k f^{(i)}(s,\cdot)\arx{\right}): 1\le k \le K, f_{1:k} \in \Fcal\arx{\right}\}.
$$
Na\"ive computation of this policy will require storing its tabular representation (i.e., $|\Scal\times\Acal|$ numbers), which is clearly inefficient in large state spaces. However,  to compute $f_{k+1} = f_{\min}^{\pi_{k+1}}$ we only need to evaluate $\pi_{k+1}$ on data points in $\Dcal$. Hence, ``lazy evaluation'' suffices, where we calculate $\pi(\cdot \mid s)$ on demand for any queried $s$. Since $\pi_{k+1}(\cdot \mid s) \propto \exp(\eta \sum_{i=1}^k f_i(s,\cdot))$, we can compute $\pi(a \mid s)$ as long as we store (the model parameters of) $f_i \in \Fcal$ for previous $i$. The normalization step will incur linear-in-$|\Acal|$ computational complexity. 

\para{Error decomposition} We now provide the analysis of PSPI, which will  explain the design of the policy update step 2b). First, 
\begin{align} \label{eq:pspi_mixture}
\sts{&~ J(\picomp) - J(\textrm{Unif}[\pi_{1:K}]) \\
= &~ 1/K \, \textstyle \sum_{k=1}^K (J(\picomp) - J(\pi_k)).}\arx{J(\picomp) - J(\textrm{Unif}[\pi_{1:K}]) = \frac{1}{K} \sum_{k=1}^K (J(\picomp) - J(\pi_k)).}
\end{align}
Now we invoke Lemma~\ref{lem:opl_error_prop} on each $k$, by choosing $f = f_k$: 
\begin{align*}
J(\picomp) - J(\pi_k) = &~ \frac{1}{1-\gamma} \Big(\EE_{s \sim d^{\picomp}}[f_k(s, \picomp) - f_k(s,\pi_k)]  \\
&~ + \EE_{d^{\picomp}}[\Tcal^{\pi_k} f_k - f_k] + \EE_{d^{\pi_k}}[f_k - \Tcal^{\pi_k} f_k] \Big).
\end{align*}
First, we apply Lemma~\ref{lem:eval_error} ``reversely'' and the 3rd term becomes
$$
\EE_{d^{\pi_k}}[f_k - \Tcal^{\pi_k} f_k ] = J_{f_k}(\pi_k) - J(\pi_k) \le 0.
$$
``$\le 0$'' follows directly from that $f_k = f_{\min}^{\pi_k}$ and $J_{f_k}(\pi_k) = J_{\textrm{VS}}^-(\pi_k)$ is a pessimistic estimation (Eq.\eqref{eq:Jvs-}). 
For the 2nd term, note that any function in $\Fcal_{\epsilon_0}^{\pi_k}$, including $f_k = f_{\min}^k(\pi_k)$, has well controlled $\EE_{D}[(f - \Tcal^{\pi_k} f)^2]$.\footnote{This requires Bellman-completeness under $\pi_k$. Given that $\pi_k$ is random, we relax it to $\Tcal^\pi f\in\Fcal, \forall f\in\Fcal, \pi\in\Pi$. Hence,  PSPI makes stronger expressivity assumptions than Theorem~\ref{thm:pessimism} (which only needs completeness for $\picomp$), and is more similar to Theorem~\ref{thm:neutral}.} This translates to the error w.r.t.~$d^{\picomp}$ under single-policy coverage, i.e., bounded $\Cavg_{\picomp}$. 

Therefore, the pessimistic design of $f_k = f_{\min}^{\pi_k}$ automatically handles the 2nd and the 3rd terms. We only need to take care of the first term now, which is 
\begin{align} \label{eq:pspi_term1}
\EE_{s \sim d^{\picomp}}[f_k(s, \picomp) - f_k(s,\pi_k)].
\end{align}
So far we have not used any properties of $\pi_k$. Hence, the only design goal for $\pi_k$, i.e., the policy update rule, is to make  Eq.\eqref{eq:pspi_term1} as negative as possible. Ideally we would like to choose $\pi_k$ to be greedily w.r.t.~$f_k$, but this is against the causal order: $f_k = f_{\min}^{\pi_k}$ is calculated \textit{given $\pi_k$ as input}! 

\para{Mirror Descent} Note that Eq.\eqref{eq:pspi_term1} is taking expectation w.r.t.~$d^{\picomp}$, so we can aggregate such terms across $k$ in Eq.\eqref{eq:pspi_mixture}: 
$$
\EE_{s \sim d^{\picomp}}\left[\frac{1}{K} \sts{\textstyle} \sum_{k=1}^K (f_k(s, \picomp) - f_k(s,\pi_k))\right].
$$
Instead of trying to come up with $\pi_k$ that maximizes $f_k(s, \cdot)$ (which is against the causal order, as noted above), we now have an easier task: design $\pi_{1:K}$ to maximize $\sum_{k} f_k(s, \cdot)$ and compete with a fixed benchmark $\picomp$. For a fixed $s$, this design problem fits a classical setting in no-regret learning: given a discrete space $\Xcal$, 
\begin{mdframed}
\textbf{Online Linear Optimization in the Simplex.} \\
For round $k=1, 2, \ldots, K$,
\begin{enumerate}[leftmargin=*]
\item Learner proposes distribution $p_k \in \Delta(\Xcal)$.
\item Nature chooses a (bounded and possibly adversarial) function $f_k \in \RR^\Xcal$. 
\end{enumerate}
Goal: minimize regret $\sum_{k=1}^K (\EE_{p}[f_k] - \EE_{p_k}[f_k])$ against any static benchmark $p\in\Delta(\Xcal)$. 
\end{mdframed}
Mapping this protocol onto our problem, we have $\Xcal \longleftarrow \Acal$, $p_k \longleftrightarrow \pi_k(\cdot \mid s)$, $f_k \in \RR^{\Xcal} \longleftrightarrow f_k(s, \cdot) \in \RR^{\Acal}$, and $p \in \Delta(\Xcal) \longleftrightarrow \picomp(\cdot \mid s) \in \Delta(\Acal)$. Therefore, classical algorithm that provides regret bound, such as \textit{mirror descent}, can be directly applied to our problem. Indeed, the policy update rule in PSPI (step 2b) is literally running mirror descent on each state individually.\footnote{This is a version of Natural Policy Gradient (NPG) \citep{kakade2001natural, agarwal2020optimality}.} With appropriate learning rate $\eta$, mirror descent enjoys the guarantee  \citep{hazan2016introduction,xie2021bellman}:
$$
\frac{1}{K}  \sum_{k=1}^K (f_k(s, \picomp) - f_k(s,\pi_k)) \lesssim\frac{\Vmax}{1-\gamma} \sqrt{\frac{\log|\Acal|}{K}}. 
$$
This shows that we can control  Eq.\eqref{eq:pspi_term1}
by increasing the number of iterations $K$. This way, we have all terms coming out of Eq.\eqref{eq:pspi_mixture} under control, under Bellman completeness and single-policy coverage.

\subsection{Pointwise Pessimism: PEVI} \label{sec:pevi}
The PSPI algorithm requires a nontrivial optimization oracle of computing $f_{\min}^\pi$. Despite its efficiency in the linear setting, its general feasibility is unclear. Therefore, it is worth asking if we can enjoy single-policy coverage under more straightforward computational oracles. 

One such example is PEVI \citep{jin2020pessimism}, which requires a regression oracle that fits $(s,a) \mapsto r + \gamma v(s')$ for any $v\in \RR^{\Scal}$ with \textbf{pointwise uncertainty quantification}. Specifically, the oracle outputs a predictor $\hat f \in \mathcal{F}$ along with $b \in \mathbb{R}^{S\times A} \ge 0$ (often called a bonus term), such that with high probability, 
\begin{align}\label{eq:pointwise-uncertainty}
f^\star(s,a) \in [\hat f(s,a) - b(s,a), \hat f(s,a) + b(s,a)],
\end{align}
where $f^\star$ be the Bayes-optimal predictor.

\para{PEVI \citep{jin2020pessimism}} Given the oracle, we can modify FQI to incorporate uncertainty $b(s,a)$ to ensure pessimism: let $f^-_0 \equiv 0$,
\begin{mdframed}
\textbf{For} $k=1, 2, \ldots, K$, 
\begin{enumerate}[leftmargin=*]
\item Fit $(s,a) \mapsto r + \max_{a'} f^-_{k-1}(s',a')$ on $\Dcal$. Let $\hat f_k \in \Fcal$  and $b_k$ be oracle outputs.
\item $f_k^-(s,a)\coloneqq \hat f_k(s,a) - b_k(s,a)$. 
\end{enumerate}
\textbf{Output:} non-stationary policy $\pi_{K:1}\coloneqq \{\pi_K, \ldots , \pi_1\}$, where $\pi_k \coloneqq \pi_{f_k^-}$, i.e., the greedy policy w.r.t.~$f_k^-$.
\end{mdframed}

\para{Error propagation and coverage definition} The first step of the analysis is to show pessimism, that $f_{k}^-(s,a) \le Q_k^{\pi_{k-1:1}}$ (recall the non-stationary setup in the FQE analysis in Section~\ref{sec:fqe}). Throughout the analysis we assume the high-probability event that all uncertainty quantification is valid. 

Pessimism can be shown by induction, based on the monotone property of Bellman operators: for any $f \le f'$ (pointwise) and $\pi$, we have $\Tcal^\pi f \le \Tcal^\pi f'$. Therefore, if we know that $f_{k-1}^- \le Q_{k-1}^{\pi_{k-2:1}}$ (the base case at $k=1$ holds trivially with $f_0^- \equiv Q_0^{(\cdot)} \equiv 0$), then 
$$
f_k^- \le \Tcal f_{k-1}^- = \Tcal^{\pi_{k-1}} f_{k-1}^- \le  \Tcal^{\pi_{k-1}} Q_{k-1}^{\pi_{k-2:1}} = Q_{k}^{\pi_{k-1:1}}. 
$$
The first inequality follows from the validity of uncertainty quantification and the Bayes-optimal predictor for the $k$-th regression is $f_k^\star = \Tcal f_{k-1}^-$; the second follows from the greediness of $\pi_k$ w.r.t.~$f_{k-1}^-$; the third from monotonicity of $\Tcal^{\pi_k}$, and the last is the finite-horizon version of Bellman equation.  Now, invoking the finite-horizon variant of Lemma~\ref{lem:opl_error_prop} (c.f.~Eq.~\eqref{eq:telescope-non-stat}) with $f_{K:1}$ playing the role of $f$, the first term on the RHS vanishes due to greediness of $\pi_{K:1}$ w.r.t.~$f_{K:1}^-$, and the third term vanishes due to pessimism. Therefore, we are only left with the second term, that 
\begin{align}
J_K(\picomp) - J_K(\pi_{K:1}) \le  \sum_{t=0}^{K-1} \gamma^t \EE_{d_t^{\picomp}}[\Tcal f_{K-t-1}^- - f_{K-t}^-].
\end{align}
We can bound the Bellman error inside expectation using uncertainty quantification again: 
$$
\Tcal f_{K-t-1}^- = f^\star_{K-t} \le \hat f_{K-t} + b_{K-t} = f_{K-t}^- + 2b_{K-t}, 
$$
which yields \citep{jin2020pessimism}
\begin{align} \label{eq:PEVI_bound}
&~ J_K(\picomp) - J_K(\pi_{K:1}) \\
\le &~\sts{\textstyle}  2 \sum_{t=0}^{K-1} \gamma^t \EE_{(s,a)\sim d_t^{\picomp}}[b_{K-t}(s,a)].
\end{align}

\para{Expressivity Assumptions and Guarantees}
The guarantee we obtain above depends on $b_k(s,a)$, i.e., the tightness of the bonus term. Next we consider concrete settings where $b_k(s,a)$ can be explicitly computed and the RHS of Eq.\eqref{eq:PEVI_bound} can be further bounded by expressions comparable to results in previous sections. 

First of all, just like FQI/FQE, we will need assumptions to ensure that each regression we solve is realizable. Note that Bellman completeness for $\Tcal$ ($\Tcal f\in\Fcal, \forall f\in\Fcal$) is \textit{insufficient} here,  because we are not backing up functions from $\Fcal$, but $f_k^- = \hat f_k - b_k$, which can be outside the function class depending on the form $b_k(s,a)$ takes. In the literature, it is often assumed that $\Tcal f\in\Fcal, \forall f\in\RR^{\Scal\times\Acal}$ to avoid the problem, as illustrated in Figure~\ref{fig:complete}.

Second, we will need pointwise confidence interval for our prediction. A canonical setting where this is feasible is linear regression: when $\Fcal$ is the linear class induced by feature map $\phi$, $\Fcal_{\phi} = \{\langle \phi, \theta \rangle: \theta\in\RR^d\}$, we can simply run (ridge) linear regression for the point estimate $\hat f_k$, with a quadratic uncertainty term:
\begin{align} \label{eq:bonus}
b_k(s,a) = \frac{\beta}{\sqrt{n}} \sqrt{\phi(s,a)^\top (\Sigma^{\textrm{ridge}}_{\Dcal})^{-1} \phi(s,a)},
\end{align}
where  $\beta$ may depend on quantities such as dimensionality $\dm$ and horizon. $\Sigma^{\textrm{ridge}}_\Dcal = \frac{1}{n}(\sum_{(s,a)\in\Dcal} \phi(s,a) \phi(s,a)^\top + I)$. 

As a remark, linear MDPs in Example~\ref{ex:low-rank} satisfies all conditions needed above. In fact, they are  \textit{equivalent}: 

\begin{proposition} 
If $\Fcal=\Fcal_{\phi}$ is the linear class induced by $\phi$ and $\Tcal f \in \Fcal, \forall f\in\RR^{\Scal\times\Acal}$, then the MDP must be a linear MDP with $\phi^\star = \phi$ as its features.\footnote{Proposition 2 of \citet{yang2019sample} claims a very similar result but replaces $\Tcal f\in\Fcal, \forall f\in \RR^{\Scal\times\Acal}$ with Bellman completeness $\Tcal f \in \Fcal, \forall f\in\Fcal$. This is incorrect, as pointed out in Proposition 3 of \citet{zanette2020exponential}: Bellman completeness + linear $\Fcal$ cannot imply linear MDP. A counterexample is bisimulation: given an arbitrarily complex transition $P$, if we let $R\equiv 0$, then aggregating all states together is a perfect bisimulation abstraction, and the corresponding linear class (see Section~\ref{sec:abs}) has dimension $|\Acal|$ and is Bellman complete \citep{chen2019information}, but $P$ is still unrestricted and may not have a low-rank factorization.}
\end{proposition}
One can compare the linear MDP setting with the ``representation learning in low-rank MDP'' setting in Example~\ref{ex:low-rank}, where $\phi^\star$ is unknown and must be learned from some class $\Phi$. PEVI does not apply to the latter setting due to the lack of linearity of the concatenated class $\bigcup_\phi \Fcal_\phi$, despite that the setting is information-theoretically tractable and can be handled by PSPI subject to the efficiency of the $f_{\min}^\pi$ oracle. That said, when there is also a class for realizing $\psi^\star$, model-based algorithms can also handle the setting and some use quadratic bonuses similar to PEVI \citep{uehara2022representation}. 

\para{Comparison of Coverage}
Plugging Eq.\eqref{eq:bonus} into Eq.\eqref{eq:PEVI_bound}, we have $J_K(\picomp) - J_K(\pi_{K:1})$ bounded by $\frac{2\beta}{(1-\gamma)\sqrt{n}}$ times the following quantity:
\begin{align*} 
\sts{\textstyle}   (1-\gamma) \sum_{t=0}^{K-1} \gamma^t \EE_{(s,a)\sim d_t^{\picomp}}[\sqrt{\phi(s,a)^\top (\Sigma^\textrm{ridge}_{\Dcal})^{-1} \phi(s,a)}].
\end{align*}
This expression plays the role of coverage in PEVI. Up to some subtle differences, 
its square is roughly 
\begin{align} \label{eq:PEVI-coverage}
\Big(\EE_{(s,a)\sim d^{\picomp}}[\sqrt{\phi(s,a)^\top \Sigma_D^{-1} \phi(s,a)}]\Big)^2,
\end{align}
which is very close to $\sigma_{\max}( \Sigma_{\pi}^{1/2} \Sigma_D^{-1} \Sigma_{\pi}^{1/2})$, the upper bound of $C_{\picomp}^{\textrm{sq}}(\Fcal_{\phi^\star})$ given in Eq.\eqref{eq:linear-sq}, Section~\ref{sec:error-prop}. In fact, if we move the square-root of Eq.\eqref{eq:PEVI-coverage} outside, the expression will become $\text{tr}(\Sigma_{\pi}^{1/2} \Sigma_D^{-1} \Sigma_{\pi}^{1/2})$, and trace is no smaller than largest eigenvalue $\sigma_{\max}$ and can be a factor of $d$ larger in the worst case. On the other hand, having square-root inside makes Eq.\eqref{eq:PEVI-coverage} smaller than $\text{tr}(\Sigma_{\pi}^{1/2} \Sigma_D^{-1} \Sigma_{\pi}^{1/2})$, so a definitive relation to Eq.\eqref{eq:linear-sq} is unclear. That said, what is clear is that PEVI's guarantee requires $\Sigma_D$ to cover all directions hit by $\Sigma_\pi$ and does not apply when $\Sigma_D$ only covers $\EE_{d^\pi}[\phi]$ (this corresponds to $\Csq_{\picomp}$ in Eq.~\eqref{eq:linear-avg}), which is a weaker condition that still enables Theorem~\ref{thm:pessimism} and PSPI's guarantee.

\subsection{Pessimism in Deep RL} \label{sec:deep}

The two algorithms, PSPI and PEVI, are both computationally efficient in linear settings. One may wonder whether they can be implemented with practical function approximation schemes, such as deep neural networks. We briefly review the theory-practice gaps below:

\para{PSPI} The problem with PSPI is that the oracle in Eq.\eqref{eq:fmin} can be difficult to implement due to its minimax nature. Generally, most algorithms that attempt at minimizing Bellman errors need minimax optimization to tackle the double sampling problem (we will see other examples in Section~\ref{sec:mis}), and there have been empirical attempts at implementing them with deep neural networks. Despite some reported successes \citep{dai2018sbeed, patterson2022generalized}, the minimax can be difficult to tune and has very different optimization properties compared to dynamic-programming (DP) algorithms; the latter have received extensive study and empirical experimentation  since the early days of deep RL \citep{mnih2015human}, and various practical tricks and improvements have contributed to their reliability and stability \citep{kumar2019stabilizing}. It is possible that further empirical research in BRM-type algorithms may enable a more faithful implementation of version-space-based pessimistic algorithms such as PSPI. 

In practice, empirical implementation of PSPI \citep{cheng2022adversarially} still uses DP/TD-style updates for value estimation.  
This, however, breaks the theoretical guarantees: in finite-horizon problems, DP is bottom-up and freezes the value functions at later time steps before fitting earlier time steps, which makes it difficult for pessimism at the initial time step to influence value estimation in later time steps. 

Another problem with PSPI is in its NPG policy update. The requirement of storing all previous iterates of value functions can be a significant burden in practice. Furthermore, the NPG step does not allow for a standalone policy class (see more discussion at the end of this subsection) and cannot scale to large action spaces computationally.  

\para{PEVI} Unlike PSPI, PEVI is designed to be a DP algorithm, and we can easily swap out linear regression for regression with neural-nets. The problem is the bonus term that performs pointwise uncertainty quantification, which can be difficult to obtain beyond the simple linear case. A common heuristic is to do regression with the neural-net, and use the last-layer feature as $\phi$ to compute the quadratic bonus \citep{bai2022pessimistic}, which again breaks the theoretical guarantee. On the other hand, PEVI can directly benefit from advances in uncertainty quantification for standard regression problems. 

\para{Behavior Regularization} Empirically, it is also common to take a standard DP algorithm and add \emph{behavior regularization}, often in the form of a regularization term $\log \pi(a \mid s)/\pi_D(a \mid s)$. This corresponds to the KL divergence between the action distributions of the learned policy $\pi$ and behavior policy $\pi_D$ \citep{fujimoto2019off}, as a way to encourage $\pi$ to stay within the offline data distribution. However, restricting $\pi(a \mid s)/\pi_D(a \mid s)$ can only control $\prod_{t=1}^H \pi(a_t \mid s_t)/\pi_D(a_t \mid s_t)$, which is the notion of coverage used by IS (Section~\ref{sec:IS}). Furthermore,  $\log(\cdot)$ is a very weak regularizer as it allows the term inside to be exponentially large, and recent works in RLHF for large language models also show that such terms alone are insufficient for preventing out-of-distribution policies (see \citet{huang2024correcting} and Section 4.2 of \citet{ye2024theoretical}). 

\para{Standalone Policy Class} Empirically, offline RL algorithms are often tested on control problems with continuous action spaces. In these problems, it is natural to have a standalone policy class $\Pi$ (instead of having it induced from $\Fcal$), often consisting of stochastic policies. While the information-theoretic algorithm of maximizing $J_{\textrm{VS}^-}(\pi)$ can handle this setting, both PSPI and PEVI crucially rely on being able to optimize the policy's action distribution separately for each state. The challenge with optimizing over a standalone $\Pi$ is that we will inevitably face trade-offs among optimization errors in different states. Roughly speaking, we will need to control $E_{d^{\picomp}}[f(s,\picomp) - f(s, \pi)]$ (Lemma~\ref{lem:opl_error_prop}) where $f$ is the pessimistic value function, but $d^{\picomp}$ is unknown. We may minimize the function under $d^D$ by $\argmax_{\pi} \EE_{d^D}[f(s,\pi)]$, but $f_k^-(s,\picomp) - f_k^-(s, \pi)$ is not a one-sided error and can be negative in some states while positive in others. As a consequence, Lemma~\ref{lem:translate} does not apply and error under $d^D$ may not translate to that under $d^{\picomp}$ even with coverage assumptions. How to handle standalone policy classes in a computationally efficient manner is still an open problem. 

\subsection{Model-based RL and Further Discussions} 
We conclude this section by briefly introducing model-based algorithms. The model-based formulation will also make it easy for further discussions on pessimism. 

\para{Model-based Algorithms} 
In RL, ``model-based'' refers to explicitly learning and using the dynamics of the MDP in the task, and by this standard all algorithms we introduced so far are model-free. Below we sketch a typical algorithmic framework for model-based offline RL, and we will see that it can be analyzed using the tools we have introduced so far. 

In theoretical analyses, the log-loss as in Maximum Likelihood Estimation\footnote{
One caveat of this formulation is that practical model-based algorithms in deep RL seldom use log-loss \citep{hafner2019dream,yu2020mopo,kidambi2020morel,bhardwaj2023adversarial}; see \citet{jiang2024note} for further discussion.} is often considered for model learning:\footnote{We assume reward function is known, as learning reward only involves a standard regression and can be easily incorporated.} given a class of candidate dynamics $\Pcal$, the loss for $\widetilde P\in\Pcal$ is
\sts{$$
\textrm{log-loss}_{\Dcal}(\widetilde P) = \textstyle 1/|\Dcal| \sum_{(s,a,s') \in \Dcal} - \log \widetilde P(s' \mid s,a).
$$}
\arx{$$
\textrm{log-loss}_{\Dcal}(\widetilde P) = \frac{1}{|\Dcal|} \sum_{(s,a,s') \in \Dcal} - \log \widetilde P(s' \mid s,a).
$$}
Concentration inequalities for MLE \citep{zhang2006eps,agarwal2020flambe} show that the minimizer of log-loss, $\Phat$, satisfies
$$
\EE_{(s,a)\sim d^D}[\|P(\cdot \mid s,a) - \Phat(\cdot \mid s,a)\|_1^2] \lesssim \frac{\log(|\Pcal|/\delta)}{n}.
$$
Now, if we output $Q_{\Phat}^\pi$, the Q-function of $\pi$ in the MDP defined by $(R, \Phat)$, we can show that it has bounded Bellman error w.r.t.~the true model: 
\begin{align*}
&~ |Q_{\Phat}^\pi(s,a) - (\Tcal^\pi Q_{\Phat}^\pi)(s,a)| \\
= &~ |(\Tcal_{\Phat}^\pi Q_{\Phat}^\pi)(s,a) - (\Tcal^\pi Q_{\Phat}^\pi)(s,a)| \\
= &~ |R(s,a) + \gamma \langle \Phat(\cdot \mid s,a), V_{\Phat}^\pi\rangle \\
 & \qquad - (R(s,a) + \gamma  \langle P(\cdot \mid s,a),  V_{\Phat}^\pi\rangle)| \\
= &~ |\gamma \langle \Phat(\cdot \mid s,a) - P(\cdot \mid s,a),  V_{\Phat}^\pi\rangle| \\
\le &~ \gamma \|\Phat(\cdot \mid s,a) - P(\cdot \mid s,a)\|_1 \Vmax. 
\end{align*}
The first step follows from that $Q_{\Phat}^\pi$ satisfies the Bellman equation in $(R, \Phat)$, and the last step follows from H\"older's inequality. Now, taking expectation over $d^D$ of the squared Bellman error and plugging in the earlier concentration result, we have
$$
\EE_{d^D}[(Q_{\Phat}^\pi - \Tcal^\pi Q_{\Phat}^\pi)^2] \lesssim \Vmax \frac{\log(|\Pcal|/\delta)}{n}.  
$$
The LHS is precisely $\Ecal(Q_{\Phat}^\pi; \pi)$, so the guarantee is exactly analogous to Eq.\eqref{eq:on-data} for BRM estimation of $Q^\pi$, and the subsequent analysis in Section~\ref{sec:brm} can be applied as-is here. 

Similarly, the information-theoretic pessimistic algorithm in Section~\ref{sec:vs-pessimism} also has its model-based counterpart. Define the version space of models as:
$$
\Pcal_{\Dcal} \coloneqq \{ \widetilde P \in \Pcal: \textrm{log-loss}_{\Dcal}(\widetilde P) \le_{\epsilon} \textrm{log-loss}_{\Dcal}(\Phat)\},
$$
where $\le_\epsilon$ is $\le$ up to a statistical threshold term similar to $\epsilon_0$ in Eq.\eqref{eq:vs}. The loss of MLE model $\Phat$ is playing a similar role to the variance correction term (Eq.\eqref{eq:g-Tf}) for value-based learning, making sure that we do not eliminate the true dynamics $P$ and only keep around candidate models whose excess risk is small on the data distribution. Then, we have the model-based analogue of $\argmax_{\pi\in\Pi} J_{\textrm{VS}}^{-}(\pi)$: 
\begin{align} \label{eq:model-based}
\argmax_{\pi\in\Pi} \min_{\widetilde P \in \Pcal_{\Dcal}} J_{\widetilde P}(\pi), 
\end{align}
which enjoys a similar guarantee to Theorem~\ref{thm:pessimism}. An interesting difference is that the model-based guarantee do not need to separately pay for $\log|\Pi|$, as the model estimation is independent of the policies.

\para{Return, Regret, and General Objectives} In most areas of RL, the following two objectives are equivalent performance measures:
$$
\textrm{Return:} ~ J(\pi) ~~\textrm{vs. ~~ Regret:} ~ J(\pi^\star) - J(\pi), 
$$
where $\pi^\star$ is the optimal policy in the MDP. 
This is because maximizing return is the same as minimizing regret, as they only differ by a negative sign and a constant shift.  A perhaps surprising fact is that they become \textit{different} when pessimism is involved: for example, consider 
$$ \sts{\textstyle} 
\argmax_{\pi\in\Pi}  \sts{\textstyle} \min_{\widetilde P \in \Pcal_D} (J_{\widetilde P}(\pi_{\widetilde P}^\star) - J_{\widetilde P}(\pi)).
$$
This will be a policy that minimizes \textit{worst-case} regret, which will be generally different from Eq.\eqref{eq:model-based} that maximizes \textit{worst-case} return; see Figure~\ref{fig:mab} for a concrete bandit example. These are design choices that we must make to reflect our preferences, and we should generally analyze an algorithm under the performance measure it uses (see e.g., \citet{bhardwaj2023adversarial}'s discussion of \citet{xiao2021optimality}). 

In fact, these are not the only choices. One can also consider a ``relative'' return  \citep{bhardwaj2023adversarial}: 
\begin{align} \label{eq:rela_pess}
J(\pi) - J(\piref), 
\end{align}
which leads to different behaviors with different $\piref$. (Note that regret is not a special case of this, because here $\piref$ is fixed but $\pi_{\widetilde P}^\star$ in the definition of worst-case regret changes with $\widetilde P \in \Pcal$.) \citet{bhardwaj2023adversarial} shows that using Eq.\eqref{eq:rela_pess} to replace return in Eq.\eqref{eq:model-based} achieves similar guarantees under the additional assumption that $\piref$ is covered. It also comes with the additional benefit that the learned policy is never worse than $\piref$ as long as $P\in \Pcal_\Dcal$, and this non-degenerate property can be desirable in industrial applications. Another benefit is that optimizing a difference can be numerically more stable than optimizing an absolute quantity. 
Furthermore, when $\piref = \pi_D$, the coverage of $\piref$ is automatically satisfied ($C_{\pi_D} = 1$ if $d^D = d^{\pi_D}$), and the algorithm also has a model-free version by transforming Eq.\eqref{eq:rela_pess} with the PD lemma (Ft~\ref{ft:PD}) \citep{cheng2022adversarially}. 

\para{Connection to Robust MDPs} The computational problem in Eq.\eqref{eq:model-based} takes the form of the robust MDP problem \citep{wiesemann2013robust}, generally written as 
$$
\argmax_{\pi\in\Pi} \min_{M \in \Mcal} J_M(\pi),
$$
where $\Mcal$ is called the uncertainty set. Robust MDP (RMDP) is concerned with approximating this problem in a computationally efficient manner, and the performance of an algorithm output $\hat\pi$ is evaluated by the gap between $\min_{M\in\Mcal} J_M(\hat\pi)$ and the optimal objective value of the robust MDP problem. Solving the problem computationally often involves computing value functions and perform (robust) Bellman updates, which are similar to the model-free pessimistic algorithms (like PEVI). 

The RMDP literature is often motivated by the fact that the true model dynamics are unknown and data-driven estimation has uncertainties, which coincides with the setting of offline RL. The difference is that RMDP assumes that the uncertainty set $\Mcal$ is given as input and focuses on the computational problem of maximizing pessimistic return, whereas offline RL takes the formation of uncertainty set as a component of the end-to-end problem and provides guarantees under coverage conditions. 

In terms of problem settings, RMDPs typically consider the tabular setting (with some recent exceptions \citep{zhou2024natural}), where the number of states is small and $\Mcal$ is factored, in the sense that the uncertainty for transition and reward is independent across the state-action space. This factorization structure is similar to the pointwise uncertainty quantification  in Section~\ref{sec:pevi}, and enables classical algorithms such as robust VI which PEVI resembles \citep{wiesemann2013robust}. 

An interesting possibility is to apply PSPI-type algorithms for RMDPs with large state spaces and arbitrary uncertainty set, when we are given the pessimistic oracle 
$M(\pi) \coloneqq \argmin_{M\in\Mcal} J_M(\pi)$, which is sometimes also considered in RMDPs.\footnote{What we really need is $Q_{M(\pi)}^\pi$, which is similar to the $f_{\min}^\pi$ oracle in Eq.\eqref{eq:fmin}.} However, a straightforward adaptation shows that PSPI only enjoys guarantees when $\Mcal$ only has reward uncertainty: let $\pirb$ be the optimal robust policy and $\hat\pi$ be the output of PSPI (the uniform mixture over iterates $\pi_{1:K}$),
\begin{align}
&~ J_{M(\pirb)}(\pirb) - J_{M(\hat\pi)}(\hat\pi) \\
= &~ J_{M(\pirb)}(\pirb) - \sts{\textstyle} \frac{1}{K}\sum_{k=1}^K J_{M(\hat\pi)}(\pi_k) \\
\le &~ \sts{\textstyle} \frac{1}{K} \sum_{k=1}^K \left( J_{M(\pi_k)}(\pirb) - J_{M(\pi_k)}(\pi_k) \right) \\
= &~ \sts{\textstyle} \frac{1}{K(1-\gamma)} \sum_{k=1}^K \EE_{d_{M(\pi_k)}^{\pirb}}[Q_{M(\pi_k)}^{\pi_k}(s, \pirb) - Q_{M(\pi_k)}^{\pi_k}(s, \pi_k)].
\end{align}
If all $M\in\Mcal$ share  the same transition dynamics, $d_{M(\pi_k)}^{\pirb}$ will be the same for all $k$, and we can push $\sum_{k=1}^K$ inside the expectation and the rest of the analysis will be the same as PSPI. However, this is not doable in more general settings with transition uncertainty.\footnote{Another way that may circumvent the issue is to assume strong coverage assumption on the initial distribution, which is a common condition in policy gradient analyses \citep{agarwal2020optimality}; c.f.~Assumption 6 in \citet{zhou2024natural} and how it is used in the proofs of their Theorems 8 and 9.} Similar difficulties are encountered when PSPI-like algorithms are applied to offline 2-player 0-sum Markov games \citep{zhang2023offline}. How to reconcile this inapplicability with the intimate relation between RMDP and offline RL, as well as how to overcome this difficulty, will be interesting directions for future investigation.

\section{Learning under Realizability and Value-Function Selection} \label{sec:select}

Value-function estimation has been a core component to all previous sections, where we learn $Q^\pi$ or $Q^\star$ from a potentially large and rich class $\Fcal$. 

Now consider a ``simpler'' version of the problem:
\begin{mdframed}
\textbf{Value-function Selection} \\
\textbf{Input: } $f_1, \ldots, f_m \in \RR^{\Scal\times\Acal}$.\\
\textbf{Output: } $f_i$ such that $f_i \approx Q^\pi$. 
\end{mdframed}
Despite its deceptive simplicity, \textbf{none of the methods so far are applicable even with $m=2$ and one of $f_1$, $f_2$ is exactly $Q^\pi$,} and algorithms like BRM require a separate helper class ($\Gcal$ in Section~\ref{sec:brm}) that realizes $\{\Tcal f_{i}\}_{i=1}^m$ to handle double sampling; the situation is similar for $Q^\star$. This means even with good data coverage (that is, we temporarily put away consideration of pessimism), we cannot perform hyperparameter tuning and model selection for FQI (e.g., selecting   neural architecture). If we use FQE to estimate the performance of trained policies, FQE's own hyperparameters will be left untuned, and so on.

The other side of the same coin is that no estimators so far work under straight  realizability, $Q^\pi \in \Fcal$; had we had such an algorithm, applying it to $\Fcal = \{f_i\}_{i=1}^m$ would solve the selection problem. Indeed,  we can  make progress on this difficult problem by switching back and forth between the two views: the \textit{learning view}, where $\Fcal$ is large and possibly structured, and the \textit{selection view}, where $\Fcal = \{f_1, \ldots, f_m\}$ is unstructured but has a small size. Studying this issue also allows us to visit several less known ideas in the literature.

\subsection{The Non-expansive Projection Argument} \label{sec:non-expand}
We start with the learning view, which asks the question: can we learn $Q^\pi$ from $\Fcal$ under realizability, without assuming Bellman completeness? To answer the question we revisit the counterexample in Proposition~\ref{prop:deadly}: when there is infinite data, each iteration of FQE  becomes
\begin{align} \label{eq:proj_bellman}
f_k \gets \proj_{\Fcal} \Tcal^\pi f_{k-1},
\end{align}
where $\proj_{\Fcal} f\coloneqq \argmin_{f'\in\Fcal} \|f - f'\|_{2, d^D}$  and corresponds to linear regression for linear $\Fcal$. To simplify presentation, in this section we assume $d^D$ is supported on the entire $\Scal\times\Acal$, so that $\proj_\Fcal$ is  unique for linear $\Fcal$.

Without the projection step, Eq.\eqref{eq:proj_bellman} is textbook value-iteration, which enjoys convergence based on the $L_\infty$ contraction of $\Tcal^\pi$:  
$$\|f_k - Q^\pi\|_\infty = \|\Tcal^\pi f_{k-1} - \Tcal^\pi Q^\pi\|_\infty \le \gamma \|f_{k-1} - Q^\pi\|_\infty.$$ 
Therefore, the divergence in Proposition~\ref{prop:deadly} can be attributed to $\proj_\Fcal$ destroying the contraction of $\Tcal^\pi$.

So a natural idea for ``fixing'' the analysis is to make sure that $\proj_\Fcal$ is \textit{non-expansive}, that for any $f, f'$, $\|\proj_\Fcal f - \proj_\Fcal f'\| \le \|f - f'\|$ for some appropriate norm $\|\cdot\|$ (see \citet{bertsekas1996neuro}, Assumption 6.3), with the hope that composing a non-expansion with a contraction still yields a contraction. In fact, $\proj_\Fcal$ \textit{is} non-expansive when $\Fcal$ is linear and $\|\cdot\| = \|\cdot\|_{2, d^D}$. Then why can the divergence in Proposition~\ref{prop:deadly} still happen to linear $\Fcal$?

The answer is that $\Tcal^\pi$ is contraction under $L_\infty$ and $\proj_\Fcal$ is non-expansion under $\|\cdot\|_{2, d^D}$, and the mismatch between norms prevents us from combining their contraction/non-expansion properties. That said, $\Tcal^\pi$ is also a contraction under weighted 2-norm, but under a special distribution: for any $f,f'\in\RR^{\Scal\times\Acal}$ and a distribution $\nu \in\Delta(\Scal\times\Acal)$ to be determined later,
\begin{align}
&~ \| \Tcal^\pi f - \Tcal^\pi f'\|_{2, \nu}^2 \\
= &~ \gamma^2 \EE_{(s,a)\sim d^D}[\big(\EE_{s'\sim P(\cdot \mid s,a)}[f(s',\pi)] \sts{\\
&~} -  \EE_{s'\sim P(\cdot \mid s,a)}[f'(s',\pi)]\big)^2] \\
\le &~ \gamma^2 \EE_{\substack{(s,a)\sim d^D, \\s' \sim P(\cdot \mid s,a), a'\sim \pi(\cdot \mid s)}}[(f - f')^2] \\
=: &~ \gamma^2 \|f - f'\|_{2, P(\nu)\times \pi}^2.
\end{align}
We start with $\nu$ but obtain a different distribution at the end, $P(\nu)\times\pi$. To make this a contraction property, let $P(\nu)\times\pi = \nu$, i.e., $\nu$ is an \textit{invariant distribution} w.r.t.~policy $\pi$. To further allow combination with $\proj_\Fcal$'s non-expansion, we need to let $d^D = \nu$; this is an ``on-policy'' assumption,\footnote{The invariant-distribution assumption here is stronger than the typical on-policy condition. To satisfy it with trajectory data (Section~\ref{sec:IS}), we need $d_0$ to be such an invariant distribution to start with. In comparison, ``on-policy'' IS (i.e., $\pi_D = \pi$ and $\CA = 1$) reduces to Monte-Carlo policy evaluation and can work with arbitrary $d_0$.} that data is sampled from the target policy we want to evaluate, beating the basic premise of  OPE.

\subsection{State Abstractions} \label{sec:abs}
To mitigate the mismatch in norms, instead of having $\Tcal^\pi$ to be a contraction under weighted 2-norm, we can instead ask $\proj_\Fcal$ to be a non-expansion under $L_\infty$ to align with the $L_\infty$ contraction of $\Tcal^\pi$. 
\citet{gordon1995stable} shows that this is indeed the case when $\Fcal$ is \textit{piecewise constant}. Let $\Fcal$ be a linear class with feature $\phi \in\RR^{\dm}$. We say $\Fcal$ is piecewise constant if $\phi$ is ``1-hot'', that $\phi(s,a)$ equals 1 in exactly one of the coordinates and 0 elsewhere. In this case, we indeed have $\|\proj_\Fcal f - \proj_\Fcal f'\|_{\infty} \le \|f - f'\|_\infty$. 

Piecewise-constant $\Fcal$ is closely related to the literature of state abstractions (or aggregations) \citep{li2006towards}.\footnote{Strictly speaking, 1-hot $\phi$ induces a partition over $\Scal\times\Acal$, which may not correspond to a well-defined partition over $\Scal$; we ignore this subtle difference in the discussion.}  A large part of this literature focuses on bisimulation, which satisfies Bellman completeness;  thus, its theoretical understanding is largely subsumed by the more general analyses we have seen in Section~\ref{sec:error-prop}. In comparison, the fact that abstractions enjoy guarantees under realizability alone\footnote{This is known as $Q^\pi$- or $Q^\star$-irrelevant abstractions \citep{li2006towards}.} is what truly makes them different from other function-approximation schemes.

The above analysis is under $L_\infty$, which is incompatible with learning in large state spaces. To have a more distribution-aware analysis similar to Section~\ref{sec:error-prop}, we define an $\phi$-aggregated MDP, $M_\phi$ with transition
\begin{align}
P_{\phi}(s' \mid s,a) = \sts{\textstyle} \sum_{\widetilde s,\widetilde a: \phi(\widetilde s,\widetilde a) = \phi(s,a)} d_D^\phi(\widetilde s,\widetilde a) P(s' \mid \widetilde s,\widetilde a).
\end{align}
and reward $R_\phi$ is defined similarly. Here $d_D^\phi(s,a) \propto d^D(s,a)$ but is normalized within a partition (that is, all $(s,a)$ pairs that share the same $\phi(s,a)$). Two key properties: (1) $Q^\pi = Q_{M_\phi}^\pi$, that $M_\phi$ preserves the true $Q^\pi$, and (2) $\proj_\Fcal \Tcal^\pi f = \Tcal_{M_\phi}^\pi f$, so projected Bellman update is essentially the true Bellman update in $M_\phi$. Then we can essentially carry out an analysis similar to Section~\ref{sec:error-prop} but in $M_\phi$.  This results in coverage assumptions that $d_D$ covers $d_{M_\phi}^\pi$, where the discounted occupancy $d^\pi$ is defined w.r.t.~the dynamics of $M_\phi$ \citep{xie2020batch, jia2024offline}. 

\subsection{BVFT} \label{sec:bvft}
The result in Section~\ref{sec:abs} removes Bellman-completeness, at the cost of imposing piecewise constant $\Fcal$, exhibiting an expressivity-structure trade-off. One can argue though that the result is not too useful for either the ``learning view'' (piecewise-constant too restrictive) or the ``selection view'' (the result does not seem to help with the selection problem). 

We now show that the benefit of strong structures can be ``lifted'' to unstructured $\Fcal$ \textit{for free}, at least information-theoretically. For this we must switch to the selection view, and consider the minimal problem posed at the beginning of Section~\ref{sec:select}: $m=2$ and $Q^\pi \in \{f_1, f_2\}$. 
Perhaps surprisingly, only using these pieces of information, we can already create a piecewise-constant $\Fcal$ with bounded dimension (i.e., the size of the partition, which determines the statistical capacity of $\Fcal$), without making further assumptions or leveraging additional side information. 

To describe the construction, it suffices to give the ``group identifier''  for each $(s,a)$, which specifies the partition that will induce $\Fcal$. To do so, we first discretize the value range $[0, \Vmax]$ to a regular grid with grid size $\epsilon$, and discretize $f_1(s,a)$ and $f_2(s,a)$ accordingly, denoted as $\bar f_1(s,a), \bar f_2(s,a)$. This pair of integer is the group identifier, i.e., $(s,a)$ and $(s',a')$ are aggregated if $\bar f_1(s,a) = \bar f_1(s',a')$ and $\bar f_2(s,a) = \bar f_2(s',a')$. 
The resulting piecewise constant $\Fcal$ satisfies $\bar f_1, \bar f_2 \in \Fcal$, which means $Q^\pi \in_{\epsilon} \Fcal$ up to a small discretization error. On the other hand, the dimensionality of $\Fcal$ (the number of groups) is $O((\Vmax/\epsilon)^2)$, independent of the size of the state-action space. 

The algorithm, known as BVFT \citep{xie2020batch},\footnote{The original work learns $Q^\star$ using a similar argument.} also extends to selecting $m$ functions via a tournament procedure of pairwise comparisons, and has been empirically tested \citep{zhang2021towards}. The error bound has $\log m$ dependence, which means a $\log|\Fcal|$ information-theoretic result for the learning setting, though the computation seems not friendly to optimization. 

\para{Hardness Results against $Q^\pi \in \Fcal$ + Bounded $C_\pi$} The analysis of BVFT inherits the coverage assumption from Section~\ref{sec:abs}, that we need bounded $d_{M_\phi}^\pi/d^D$ for all partitions $\phi$ created in the pairwise process, which is a complicated assumption. Ideally, the cleanest assumptions would be $Q^\pi \in\Fcal$ and the standard coverage parameter $C_\pi$. The $Q^\star$ counterpart of this result was conjectured to be impossible \citep{chen2019information}, though subsequent negative results were w.r.t.~different and often weaker data assumptions \citep{wang2020statistical, amortila2020variant, zanette2020exponential}. The gap was eventually closed by \citet{foster2021statistical} (see also \citet{jia2024offline}) which confirmed the earlier conjecture, that accurate estimation of $J(\pi)$ under $Q^\pi \in \Fcal$ and bounded $C_\pi$ is information-theoretically intractable.

\subsection{LSTDQ} \label{sec:lstd}
Besides state abstractions, there is another setting where learning $Q^\pi$ only requires realizability: the LSTD algorithms. LSTD is derived by taking a TD-style algorithm with linear function approximation and directly writing down its fixed point. For example, if we take FQE with a linear $\Fcal$, its fixed point, LSTDQ, is $f_\theta(s,a)  = \phi(s,a)^\top \theta$, with $\theta = A^{-1} B$ and
$$
A = \Sigma_D - \gamma \EE_D[\phi(s,a)\phi(s',\pi)^\top], B = \EE_D[\phi(s,a)\cdot r].
$$
Here $A$ and $B$ are both population statistics, and the actual algorithm uses their finite-sample estimation $\widehat A$, $\widehat B$ to compute $\widehat \theta$. Note that the algorithm is only well-defined when $A$ is invertible, and numerically stable when $\sigma_{\min}(A)$ is bounded away from $0$ (which guarantees boundedness of $\hat \theta$); here $\sigma_{\min}$ is the smallest singular value since $A$ is not necessarily symmetric. Despite that LSTD is inspired by TD algorithms, there are cases where TD diverges but LSTD outputs bounded solutions, and LSTD algorithms generally work under more lenient conditions \citep{perdomo2023complete}. 

Interestingly, many algorithms, such as BRM (Section~\ref{sec:brm}) and MIS (Section~\ref{sec:mis}), reduce to LSTDQ when using linear classes \citep{antos2008learning, uehara2019minimax, xie2021bellman}. Hence, LSTDQ can also be analyzed under the assumptions made in those sections, which require additional expressivity assumptions other than $Q^\pi \in \Fcal$. 

But this is not necessary, and the simple linear-algebraic structure of LSTDQ admits an alternative condition.  It turns out that all we need is $\sigma_{\min}(A)$! The idea is simple: (1) the linear coefficient of $Q^\pi$ (i.e., $Q^\pi = \phi(s,a)^\top \theta^\pi$) is a solution to $A \theta = B$, and (2) if $A$ is invertible ($\sigma_{\min}(A)>0$), the solution is unique. In the finite-sample case, we can use standard concentration arguments to bound $\|A - \widehat A\|_2$ (operator norm), $\|B - \widehat B\|_2$, and 
\begin{align}
&~ \|\theta^\pi - \widehat \theta\|_2 = \|A^{-1} A \theta^\pi - A^{-1} A \widehat \theta\|_2 \\
= &~ \|A^{-1} B - A^{-1} A \widehat \theta\|_2 
\le \|B - A \widehat \theta\|_2/\sigma_{\min}(A) \\
=  &~\|B - A \widehat \theta + \widehat A \widehat \theta - \widehat B \|_2/\sigma_{\min}(A) \\
\le &~ (\| B - \widehat B\|_2 + \| A  - \widehat A \|_2 \|\widehat \theta\|_2)/\sigma_{\min}(A).
\end{align}

\para{Coverage in LSTDQ} All guarantees we have seen so far rely on three types of conditions: (1) expressivity of function class, (2) structure of function class, and (3) coverage of data. For LSTDQ, $Q^\pi \in \Fcal$ is the expressivity condition, linear $\Fcal$ is the structural condition, so the remaining $\sigma_{\min}(A)$ must correspond to coverage. Indeed, one can show that in the strictly on-policy case, that is, when $d^D$ is an invariant distribution w.r.t.~$\pi$ (Section~\ref{sec:non-expand}), $\sigma_{\min}(A)$ is positive and controlled by $(1-\gamma)$. Roughly speaking, this is because $\phi(s',\pi)$ has the same marginal distribution as $\phi(s,a)$,\footnote{For stochastic policy we can replace $\phi(s',\pi)$ in the definition of $A$ with $\phi(s',a')$, $a'\sim \pi(\cdot \mid s')$, and this claim will hold.} so $\Sigma_D = \EE_D[\phi(s,a) \phi(s,a)^\top]$ dominates the spectrum of $\EE_D[\phi(s,a)\phi(s',\pi)^\top]$. See \citet{lazaric2012finite, pires2012statistical} for more detailed analyses of LSTD in the on-policy case.

Unfortunately, our understanding of $\sigma_{\min}(A)$ outside the on-policy setting is very limited, and the parameter itself is not a very clean coverage parameter and mixes together many other factors. For example, $\sigma_{\min}(A)$ can be $0$ if there are simply redundant features in $\phi$. Another problem is that $\sigma_{\min}(A)$ is not scale-invariant: if we scale $\phi$ by a constant (and scale $\theta^\pi$ accordingly), $\sigma_{\min}(A)$ will also grow or shrink superficially. Even if we fix these issues, there is still very limited interpretability of the condition outside the on-policy case. This is a general theme with algorithms that only require realizability, that error propagation and coverage are conceptually much more complicated and hard to  interpret. 

As a final remark, just as BVFT ``lifts'' the benefit of state abstractions to unstructured function classes, a similar procedure can also lift LSTDQ, which produces an algorithm that is simpler than BVFT for learning $Q^\pi$ \citep{liu2025model}.  

\section{Marginalized Importance Sampling} \label{sec:mis}
Density ratio $d^\pi/d^D$ has played an important role in the analyses of previous sections, though they have never directly appeared in the algorithms. In this section we show that they can also be explicitly modeled (and sometimes learned) by function approximation, which leads to new algorithms for OPE or even policy optimization. The density ratio $d^\pi/d^D$ is often modeled together with value functions with an intriguing symmetry: we can learn value functions using density ratios as ``discriminators'', or vice versa. 
These algorithms, generally known as ``marginalized importance sampling'' (MIS), require different assumptions than the ones introduced previously and have complementary properties. 

\subsection{OPE via Value Functions}
We start with learning value functions for OPE. Previous guarantees for value-based OPE all require Bellman completeness. Below we show an alternative algorithm and analyses that only require realizability. In addition to $Q^\pi \in \Fcal$, we will also require a \textit{weight function class} $\Wcal$ such that $w^\pi \in \Wcal$, where $w^\pi(s,a) = d^\pi(s,a)/d^D(s,a)$. (These assumptions will be relaxed later, but for now we use them for a clean derivation.) It starts from Lemma~\ref{lem:eval_error} and only takes a few steps \citep{uehara2019minimax, jiang2020minimax}:
\begin{align}
&~ |J_f(\pi) - J(\pi)| = \left|\frac{1}{1-\gamma}\EE_{d^\pi}[f - \Tcal^\pi f]\right| \\
=  &~ \left|\frac{1}{1-\gamma}\EE_{d^D}[\nicefrac{d^\pi}{d^D} \cdot (f - \Tcal^\pi f)]\right| \\
\le &~ \max_{w\in\Wcal} \left|\frac{1}{1-\gamma}\EE_{d^D}[w \cdot (f - \Tcal^\pi f)]\right|  
=:   \max_{w\in\Wcal} L_q(w, f). 
\end{align}
The MQL algorithm minimizes (the estimation of) the above loss over $f\in\Fcal$ and estimate $J(\pi) \approx J_f(\pi)$ for the learned $f$. Here, since the ultimate goal is to make sure $J_f(\pi) \approx J(\pi)$, the derivation starts with their difference and hope to explicitly control it. Lemma~\ref{lem:eval_error} tells us that the difference is the \textit{average Bellman error} under $d^\pi$; there are two key properties of this quantity $\EE_{d^\pi}[f - \Tcal^\pi f]$:
\begin{enumerate}[leftmargin=*]
\item $f-\Tcal^\pi f$ can be positive or negative, and the positive/negative errors can cancel with each other in a plain expectation $\EE_{(\cdot)}[f  - \Tcal^\pi f] = 0$.
\item $d^\pi$ is an unknown distribution. 
\end{enumerate}
Previous algorithms, such as BRM, handle these challenges by squaring the Bellman error, which makes it one-sided (non-negative), so that controlling the (squared) error on $d^D$ indirectly controls it on any covered distribution, which also solves the problem of unknown $d^\pi$. The cost is the double-sampling issue and the introduction of the completeness assumption.

In comparison, MIS handles it very differently. It does not square the error, and $L_q(w,f)$ is directly amendable to statistical estimation and free from the double-sampling issue: $L_q(w,f) = $
$$
\left|\frac{1}{1-\gamma}\EE_{(s,a)\sim d^D}[w(s,a) (f(s,a) - r - \gamma f(s',\pi))]\right|.
$$
The problem is that we cannot just minimize $|\EE_{d^D}[f - \Tcal^\pi f]|$, as the lack of non-negativity does not allow error bound to translate  from $d^D$ to a covered distribution. The solution, as Lemma~\ref{lem:eval_error} suggests, is to use importance weights $d^\pi / d^D$ to reweight the data distribution to be precisely $d^\pi$. Of course, $d^\pi$ is unknown, so we reweight using a rich class of weight functions $w\in\Wcal$, and take the worst-case reweighted error among them. The extra functions in $\Wcal$ does not hurt, in the sense that $f = Q^\pi$ satisfies $f - \Tcal^\pi f \equiv 0$ so $L_q(w, Q^\pi) \equiv 0$, $\forall w$. 

A more formal guarantee is as follows:
\begin{theorem}[Guarantee of MQL \citep{uehara2019minimax}] \label{thm:mql}
Assume $w^\pi \in \Wcal$ and $Q^\pi \in \Fcal$. 
Let $\hat f\coloneqq \argmin_{f\in\Fcal} \max_{w\in\Wcal} \widehat L_q(w,f)$, where $\widehat L_q(\cdot)$ is the empirical estimation of $L_q(\cdot)$. Then, w.p.~$\ge 1-\delta$,
$$
|J_{\hat f}(\pi) - J(\pi)| \lesssim \frac{\Vmax \|\Wcal\|_\infty}{1-\gamma} \sqrt{\frac{\log (|\Fcal||\Wcal|/\delta)}{n}}, 
$$
where $\|\Wcal\|_\infty \coloneqq \max_{w\in\Wcal} \|w\|_\infty$. 
\end{theorem}
The guarantee is similar to Eq.\eqref{eq:antos_error} for BRM, except that the coverage parameter $C_\pi$ seems missing from the bound. The reality is that $C_\pi$ is hidden in $\|\Wcal\|_\infty$, the boundedness of the $\Wcal$ class, as $C_\pi = \|w^\pi\|_\infty \le \|\Wcal\|_\infty$. In fact, $\|\Wcal\|_\infty$ will be generally greater than $C_\pi$ if it includes $w$ with $\|w\|_\infty > \|w^\pi\|_\infty$. 

For now let's treat $\|\Wcal\|_\infty \approx C_\pi$. Another difference is that Eq.\eqref{eq:antos_error} scales with $\sqrt{C_\pi}$, but Theorem~\ref{thm:mql} scales linearly with $C_\pi$. This is a looseness   and can be tightened with some additional assumptions/pre-processing. Since $w\in\Wcal$ models the importance weights, we can assume that $\EE_{d^D}[w] \approx 1$,\footnote{If not we can replace $w$ with $w/\EE_{\Dcal}[w]$. See also Footnote~\ref{ft:is-var}.} in which case $\EE_{d^D}[w^2] \le \|w\|_\infty$ and Bernstein's inequality can give the square-root improvement just as in the analysis of IS . 

\para{Do We Really Need $w^\pi\in\Wcal$?} A somewhat common belief is that MIS has a disadvantage that it requires boundedness of $C_\pi$ as its coverage parameter and cannot enjoy refined version such as $\Csq_\pi$ or $\Cavg_\pi$. On a related note, in structural models such as linear MDPs (Example~\ref{ex:low-rank}), MIS is believed to be not applicable without a separate $\Wcal$ class as we do not know the form of $w^\pi$ and it may not be linear in $\phi^\star$. 

This is actually not true. Inspecting the derivations, we see that what we really need is a function $w^\pi_{\eff}$ such that
$$
\EE_{d^\pi}[f - \Tcal^\pi f] = \EE_{d^D}[w_\eff^\pi (f - \Tcal^\pi f)].
$$
When $f - \Tcal^\pi f$ is linear in some features $\phi$ (the algorithm does not need to know $\phi$, so this is not imposing Bellman-completeness on $\Fcal$), a sufficient condition is that \citep{zhang2024curses}
$$
\EE_{d^\pi}[\phi]= \EE_{d^D}[w_\eff^\pi \cdot \phi].
$$
This is known as the \textit{(kernel) mean matching} problem \citep{gretton2008covariate}. While $w_\eff^\pi = w^\pi$ is a solution, it can have better-behaved solutions, such as
$$
w_{\eff}^\pi(s,a) = \phi(s,a)^\top \Sigma_D^{-1} \EE_{d^\pi}[\phi].
$$
Interestingly, the 2nd moment of $w_{\eff}^\pi$ (which controls the $1/\sqrt{n}$ term in Bernstein's inequality) is $\EE_{d^D}[(w_{\eff}^\pi)^2] = \Cavg_\pi$! Furthermore, $w_{\eff}^\pi$ constructed above is linear in $\phi(s,a)$, which makes it directly applicable in linear MDPs. The slight disadvantage of MIS is that we also pay $\|w_{\eff}^\pi\|_\infty$ in the $1/n$ term in Bernstein's, which can be mitigated by using median-of-means estimators \citep{lerasle2019lecture}. 

As a final remark, due to the convexity of $L_q(w, f)$ in $w$, $w^\pi$ (or $w_\eff^\pi$) $\in\Wcal$ can be relaxed to $w^\pi \in \textrm{conv}(\Wcal)$, where $\textrm{conv}(\cdot)$ is the convex hull \citep{uehara2019minimax}. 

\para{Computational Efficiency} Similar to BRM, the MIS estimator also requires minimax optimization. When the $\Wcal$ class, which plays the role of a ``discriminator'', is an RKHS, one can show that the gradient of $\max_{w\in\Wcal} L_q(w,f)^2$ w.r.t.~$f$'s parameters has a closed-form solution\footnote{The square in $L_q(w,f)^2$ is outside the expectation and absolute value, so it does not change the statistical properties of the algorithm.} \citep{liu2018breaking, feng2019kernel}, and the objective can be effectively optimized by a single SGD over $f$. However, if we want to leverage neural-nets for $\Wcal$, the optimization can become difficult and this contributes to MIS mehtods' lack of empirical popularity. 

\para{Uncertainty Quantification} We can also derive valid upper and lower bounds on $J(\pi)$ similar to $J_{\textrm{vs}}^-(\pi)$ in Eq.\eqref{eq:Jvs-}. All we need to do is to replace the loss $\Ehat(f;\pi)$ with $\max_{w\in\Wcal} L_q(w,f)$ and set the appropriate statistical threshold \citep{jiang2020minimax, feng2020accountable}. As long as $Q^\pi\in\Fcal$, $\min J_f(\pi)$ over the version space is always a true lower bound, and it will be tight if we further have $w^\pi \in \Wcal$. This immediately leads to pessimistic policy optimization algorithms that parallel those introduced in Sections~\ref{sec:pess} and \ref{sec:pspi} \citep{jiang2020minimax,zhu2023importance}. For example, information-theoretic MIS algorithm can achieve a similar guarantee to Theorem~\ref{thm:pessimism}, under the assumptions that $Q^\pi \in \Fcal, \forall \pi\in\Pi$, and $w^{\picomp} \in \Wcal$ \citep{jiang2020minimax}. 

\subsection{OPE via Weight Functions}
MQL learns a value function using marginalized importance weights $w$ as ``discriminators''. However, their roles can be swapped: we can learn $w^\pi$ using $\Fcal$ as discriminators. In fact, MIS was first introduced in this form \citet{liu2018breaking, xie2019towards} and the discovery of value learning methods and unification happened later \citep{uehara2019minimax, feng2019kernel, jiang2020minimax}. 

The derivation of weight learning methods is similar. First, once we obtain $w^\pi = d^\pi/d^D$, we can estimate return as $J(\pi) = \EE_{(s,a)\sim d^\pi}[R(s,a)] /(1-\gamma)= \EE_{d^D}[w^\pi \cdot r] /(1-\gamma) =: J_{w^\pi}(\pi)$, with a slight abuse of notation that $J_f(\pi)$ and $J_w(\pi)$ correspond to different expressions that depend on the nature of the subscript. Similar to Lemma~\ref{lem:eval_error}, we now need a error decomposition lemma that translates the error of $J(\pi) - J_w(\pi)$ for an arbitrary $w$ into some kind of one-step Bellman-like error:
\begin{lemma}\label{lem:bellman-flow}
Given any $w\in \RR^{\Scal\times\Acal}$,  $J(\pi) - J_w(\pi) =$
$$
\EE_{s\sim d_0}[Q^\pi(s,\pi)] + \EE_{D}[w(\gamma Q^\pi(s',\pi) - Q^\pi(s,a))] / (1-\gamma).
$$
\end{lemma}
Just as the RHS of Lemma~\ref{lem:eval_error} contains the Bellman error w.r.t.~$\Tcal^\pi$ (which defines $Q^\pi$), the RHS of Lemma~\ref{lem:bellman-flow} is the violation of Bellman \textit{flow} equation for $d^\pi$:\footnote{Interestingly, Lemma~\ref{lem:eval_error}'s proof uses the Bellman flow equation for $d^\pi$, and Lemma~\ref{lem:bellman-flow}'s proof uses the Bellman equation for $Q^\pi$.} $d^\pi(s,a) = (1-\gamma) d_0(s, \pi) + \gamma \sum_{s',a'} d^\pi(s',a') P(s \mid s',a') \pi(a \mid s)$. In fact, when $w = w^\pi$, the RHS will always be $0$ even if we replace $Q^\pi$ with any other function $f$, which leads to the following  loss: $L_w(w,f)\coloneqq $
$$
\left|\EE_{s\sim d_0}[f(s,\pi)] + \EE_{D}[w \cdot (\gamma f(s',\pi) - f(s,a))] / (1-\gamma)\right|. 
$$
The MWL algorithm, 
$\argmin_{w\in\Wcal} \max_{f\in\Fcal}  \widehat L_w(w,f)$, has similar and symmetric properties w.r.t.~MQL. For  unification between the two class of methods and their duality, we refer the readers to \citet{jiang2020minimax,nachum2020reinforcement}  for further reading.

\para{Squared-loss Algorithms for Learning $w^\pi$} We have seen algorithms that learn $Q^\pi$ under Bellman completeness (FQE and BRM), and algorithms that model $Q^\pi$ and $w^\pi$ jointly (MIS). Naturally, there are also algorithms that learn $w^\pi$ under a form of completeness w.r.t.~the Bellman flow operator, both in DP style \citep{huang2023reinforcement} and BRM style \citep{nachum2019dualdice}. As notable difference between learning $Q^\pi$ and $w^\pi$ under the respective completeness assumptions, the existence of $Q^\pi$ does not depend on the property of the offline data distribution $d^D$, but the existence of $w^\pi$ does, so sometimes the learning target does not even exist for $w^\pi$ estimation when the data lacks sufficient coverage. To handle this problem, \citep{huang2023reinforcement} shows that for an arbitrary data distribution, one can always estimate a clipped version of $w^\pi$, which can also be used for pessimistic evaluation. Another benefit of learning $w^\pi$ directly is that it allows the plug-in use of more general objective functions (and constraints) of the occupancy measure  $d^\pi$, for which  the expected return $J(\pi) = \EE_{d^\pi}[R]/(1-\gamma)$ is a special case (linear objective) \citep{zahavy2021reward, mutti2023convex}. 

\subsection{Policy Optimization}
The symmetry between value-function and weight learning is closely related to the Linear Programming (LP) duality \citep{puterman2014markov}. In fact, it is well known that finding the optimal policy in an MDP can also be cast as an LP, where the saddle point of the primal-dual form corresponds to $d^{\pi^\star}$ and $V^\star(s) = Q^\star(s, \pi^\star)$, respectively, from which the optimal policy $\pi^\star$ can be extracted. 
This observation leads to the hope that instead of evaluating each policy and choosing the best among them, which requires assumptions like $Q^\pi \in\Fcal, \forall \pi\in\Pi$, the LP for $\pi^\star$ may enable an MIS algorithm that directly solves $(d^{\pi^\star}, V^\star)$ without policy evaluation, and provide single-policy coverage guarantee under only two realizability assumptions: $d^{\pi^\star}/d^D \in \Wcal$, $V^\star \in \Vcal$. This is roughly what is proved by \citet{zhan2022offline}, except that strong regularization needs to be added to prevent degenerate solution in the function-approximation setting, which leads to a relatively slow rate of $1/\epsilon^6$ for the optimality guarantee.  Improvement of the rate often comes with the cost of additional assumptions \citep{chen2022offline, ozdaglar2023revisiting,zhu2023importance}.

\section{Emerging directions and Discussions} \label{sec:discuss}
We conclude the article by discussing emerging directions and challenges. 

\para{Connection to Online RL}
Online RL, which emphasizes efficient exploration and active data collection, is another core area of RL. Despite its largely parallel development to offline RL, more and more similarities and connections have been found. To start with, the function-approximation assumptions induced in this article are also highly relevant to online RL \citep{jiang2017contextual, jin2021bellman}. 
Moreover, the central challenge in online RL is to \textit{explore}, i.e., finding policies that visit states and actions \textit{not} covered by existing data, which is often achieved by \textit{optimistic} algorithms \citep{auer2002finite}. This exhibits an interesting symmetry with pessimism in offline RL. Indeed, the key behind both optimism and pessimism is uncertainty quantification, and both version-space and pointwise pessimism have their optimistic counterparts in online RL \citep{jiang2017contextual, jin2021bellman, jin2020provably}. 

That said, the coverage conditions extensively discussed in this article seem a purely offline concept, since it depends on an offline data distribution which does not exist in the online setting. 
In online RL, what is needed is often structural assumptions on the dynamics (e.g., low-rankness in Example~\ref{ex:low-rank}) \citep{jiang2017contextual}. While structural dynamics and coverage may seem unrelated, we have seen in Section~\ref{sec:neutral_po} that the \textit{existence} of $d^D$ that allows all-policy coverage (low $\max_{\pi\in\Pi} C_\pi$) may imply restriction on the environment dynamics, leading to the suspicion that such a condition may be useful for online exploration. Indeed, recent work of \citet{xie2022role} shows that \textit{coverability}, defined as $\inf_{d^D} \max_{\pi\in\Pi} C_{\pi}$, enables sample-efficient online exploration under Bellman completeness under $\Tcal$. The quantity is inspired by offline RL coverage, but is purely a structural property of the MDP dynamics and the policy class (note the $\inf_{d^D}$). 

Besides connections on theoretical understanding and learnability conditions, there is also significant interest in exploring more ``hybrid'' learning protocols between online and offline RL, e.g., when online RL can benefit from some offline data \citep{xie2021policy,wagenmaker2023leveraging,song2022hybrid}, and when offline RL can benefit from additional online experimentation budget \citep{huang2023non}. 

\para{Intersection with Deep Learning Theory} Throughout this article we have been using na\"ive complexity measures for in-distribution  generalization error bounds, the log cardinality of finite classes. While many analyses can be extended to handle infinite classes with bounded covering numbers, 
there are also significant challenges in incorporating more modern complexity measures, especially for deep neural networks. For example, when analyzing DP-based algorithms (e.g., FQI or its online variant, such as DQN \citep{mnih2015human}), we need to deal with the loss $\Lhat(f';f,\pi)$. Fixing $f$ and $\pi$, minimizing the loss is a standard regression problem, and analyses from deep learning theory can be directly adopted. However, unlike standard supervised learning where the labels are independent of each other, here the regression label is $r + \gamma f(s',\pi)$, where $f$ and $\pi$ may be the outcome of previous iterations (e.g., $f = f_{k-1}$ in FQE/FQI) which depends on the entire dataset. Replacing union bound over $f\in\Fcal, \pi\in\Pi$ with proper complexity measures has proved difficult. These issues are often circumvented by using fresh samples in each iteration \citep{fan2020theoretical}, so that the labels can be viewed as independent for the current regression. 
\citep{antos2008learning} has addressed a related problem by proposing a new complexity measure called VC-crossing dimension, but the definition is somewhat restrictive and has not seen further development.

\para{Multi-agent Setting} Multi-agent RL (MARL) is an extension of standard (single-agent) RL, where there are multiple agents acting simultaneously with possibly misaligned or even competitive objectives.\footnote{A large part of empirical MARL   is concerned with fully cooperative settings with emphases on decentralized training and/or execution, which is a very different area; see \citet{amato2024partial} for a survey.} Central to the setting are game-theoretic concepts such as equilibria. 
A perhaps surprising fact is that even for zero-sum two-player games, coverage of the equilibrium policy $(\mu^\star, \nu^\star)$ (which are policies for each player/agent, respectively) is insufficient for offline learning. A stronger and sufficient condition is called unilateral coverage, where data covers $(\mu^\star, \nu)$ and $(\mu, \nu^\star)$ for all  $\mu$ and $\nu$ in each player's policy class \citep{cui2022offline, cui2022provably}. 

Information-theoretically,  algorithms with guarantees in the general function-approximation settings have also been developed, which build on and extend the ideas behind version-space pessimism (Theorem~\ref{thm:pessimism}). To find a (say) Nash equilibrium, a formulation amendable to learning and optimization is to minimize the equilibrium gap, e.g., $\textrm{Gap}(\mu,\nu) = \max_{\mu^\dagger} J(\mu^\dagger, \nu) - \min_{\nu^\dagger} J(\mu, \nu^\dagger)$, which measures how much the players/agents want to deviate from a candidate solution. \citep{zhang2023offline} shows that we can have a conservative estimate of the gap:  
$$ \textrm{Gap}(\mu,\nu) \le \max_{\mu^\dagger} J_{\textrm{VS}}^+(\mu^\dagger, \nu) - \min_{\nu^\dagger} J_{\textrm{VS}}^-(\mu, \nu^\dagger),$$
where $J_{\textrm{VS}}^+$ is similar to $J_{\textrm{VS}}^-$ but we look for the most optimistic estimate. The estimation of these upper and lower bounds is no different from Section~\ref{sec:pess}, since once the players' policies are fixed, the game-theoretic aspects simply disappear in the policy evaluation subproblem. Minimizing such a upper bound on the gap leads to provable guarantees under unilateral coverage.

\para{Partial Observability} Another extension of the standard MDP framework is to consider partial observability (non-Markovianity), often modelled as Partially Observable MDPs (POMDPs). POMDPs and the related PSRs formulations have been studied in the literature, but mostly from a computational perspective and often in the tabular setting. To handle large observation spaces, one idea is to view POMDPs as MDPs with history of observations and actions as the state representation; this perspective makes MDP algorithms and analyses directly applicable under appropriate function approximation over histories. However, the notion of coverage, especially that based on state density ratio $C_\pi$, becomes the probability ratio on \textit{histories}, which is the exactly the cumulative importance weights of IS (Section~\ref{sec:IS}), thus erasing the advantage of most approaches introduced in this article compared to IS. Obtaining nontrivial OPE guarantees under general function approximation is a challenging task and the investigation has started relatively recently \citep{uehara2022future, zhang2024curses}, revealing that potentially different coverage conditions are needed for partially observable environments (c.f.~belief and outcome coverage in \citet{zhang2024curses}). 

The POMDP formulation can also be used to model data confoundedness, an important consideration in causal inference from observational data \citep{tennenholtz2020off, kallus2020confounding, shi2022minimax, bruns2021model, bruns2023robust, bennett2024proximal}. A typical setting is that the behavior policy can  depend on the latent state which is not logged in the data. OPE in such a \textit{confounded} POMDP setting is related to but also very different from the previous (unconfounded) POMDP setting, and the existence of latent confounders bring significant challenges and require technical tools from the causal inference literature.

\bibliographystyle{plainnat}
\bibliography{RL}

\end{document}